\documentclass[twoside,leqno,twocolumn]{article}
\usepackage{ltexpprt}
\usepackage[pdftex]{graphicx}
\usepackage[font=footnotesize]{subfig}
\usepackage{amsmath}
\usepackage{amssymb}
\usepackage{wrapfig}

\DeclareMathOperator*{\argmin}{arg\,min}
\usepackage{color}
\newcommand{\orbi}{\textbf{ORBIT-S}}
\newcommand{\orbt}{\textbf{ORBIT-T}}
\newcommand{\orbst}{\textbf{ORBIT-ST}}
\newcommand{\orbc}{\textbf{ORBCOR}}
\begin{document}

\title{\Large ORBIT: Ordering Based Information Transfer Across Space and Time \\ for Global Surface Water Monitoring}
\author{Ankush Khandelwal\thanks{khand035@umn.edu, University of Minnesota} \\
\and
Anuj Karpatne\thanks{karpa009@umn.edu, University of Minnesota}
\and
Vipin Kumar\thanks{kumar001@umn.edu, University of Minnesota}}
\date{}

\maketitle







\begin{abstract} \small\baselineskip=9pt
Many earth science applications require data at both high spatial and temporal resolution for effective monitoring of various ecosystem resources. Due to practical limitations in sensor design, there is often a trade-off in different resolutions of spatio-temporal datasets and hence a single sensor alone cannot provide the required information. Various data fusion methods have been proposed in the literature that mainly rely on individual timesteps when both datasets are available to learn a mapping between features values at different resolutions using local relationships between pixels. Earth observation data is often plagued with spatially and temporally correlated noise, outliers and missing data due to atmospheric disturbances which pose a challenge in learning the mapping from a local neighborhood at individual timesteps. In this paper, we aim to exploit time-independent global relationships between pixels for robust transfer of information across different scales. Specifically, we propose a new framework, ORBIT (Ordering Based Information Transfer) that uses relative ordering constraint among pixels to transfer information across both time and scales. The effectiveness of the framework is demonstrated for global surface water monitoring using both synthetic and real-world datasets.\end{abstract}

\section{Introduction}
\begin{figure}[!t]%
    \centering
    \subfloat[Sep 13, 2004 from MODIS product]{\includegraphics[width=0.2\textwidth]{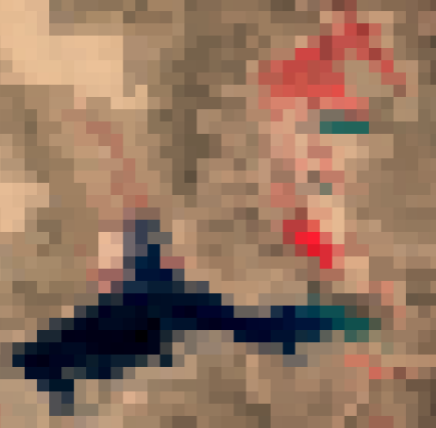}}
    \hfill
    \subfloat[Sep 16, 2004 from Landsat 7]{\includegraphics[width=0.2\textwidth]{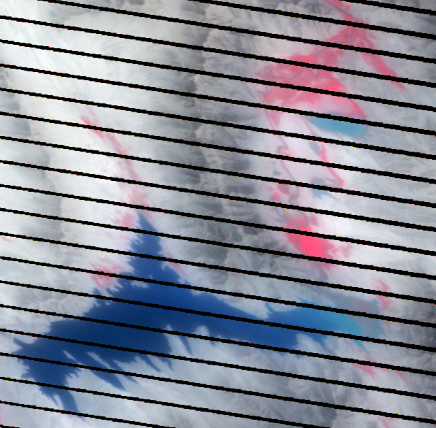}}
    \caption{An illustrative example showing False Color Composite images from two different sensors onboard Earth Observation satellites.}%
    \label{fig:resdif}%
    \vspace{-0.2in}
\end{figure}

In many applications involving spatio-temporal phenomena, data sets are available at multiple spatial and temporal scale. Often times, due to physical limitations in sensor design and cost issues, there is a trade-off in different resolutions of these datasets. As an example, consider the Earth Observation (EO) data acquired through various remote sensing satellites. MODIS sensor onboard TERRA and AQUA satellites capture earth's surface every day at a coarse spatial resolution (500m). On the other hand ETM+ sensor onboard LANDSAT 7 satellite captures the earth's surface every 16 days but at a high spatial resolution of 30m (see Figure \ref{fig:resdif}). Thus, a single sensor is not enough to provide both high spatial and temporal detail required in many earth science applications.
Hence, there is a need to develop methods that can transfer information across scales as well as across time to effectively use the rich complementary information available in these datasets.
Various methods have been proposed that aim to learn a mapping between low spatial resolution (LSR) instances and high spatial resolution (HSR) instances (exploiting local relationships between instances) at time-steps when both datasets are available. Such a mapping can then be used to generate the data at HSR data for time steps when only LSR data is available. These methods in general have limited performance because: (a) a single snapshot in time can have large amount of noise and missing data and hence might not have enough information to learn robust mapping between two resolutions, and (b) the transfer of information across scales is possible only in the duration when both datasets are available. In this paper, we propose a framework that aims to exploit inherent \emph{time-independent} relationships that exist between \textit{all instances} (instead of using local relationships) to transfer information across these complementary datasets. The use of these high-level time-independent relationships not only make the method more robust to noise and missing values but also allows for information transfer across datasets available from different duration (without necessarily having overlapping time steps). 

We demonstrate the utility of the proposed approach for the application of monitoring surface water extent variations from the Earth Observation data.

\textbf{Importance of Global Surface Water Monitoring:} Fresh water, which is only available in inland water bodies such as lakes, reservoirs, and rivers, is increasingly becoming unevenly distributed across the world, a situation which is posing a threat to human sustainability. Monitoring the dynamics of inland water bodies at a global scale has become important for: (a) devising effective water management strategies, (b) assessing the impact of human actions on water security, (c) understanding the interplay between the spatio-temporal dynamics of surface water and climate change, and (d) near-real time mitigation and management of disaster events such as floods.

\begin{figure}[!t]%
    \centering
    \subfloat[False Color Composite]{\includegraphics[width=0.23\textwidth]{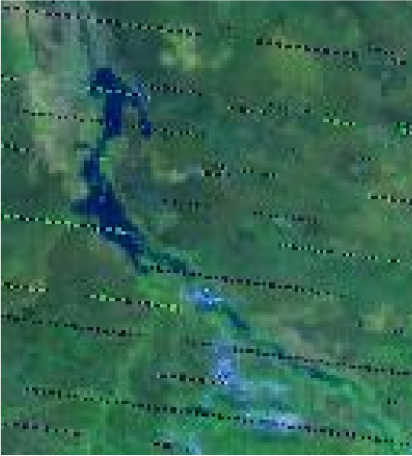}}
    \hfill
    \subfloat[Classification Map]{\includegraphics[width=0.23\textwidth]{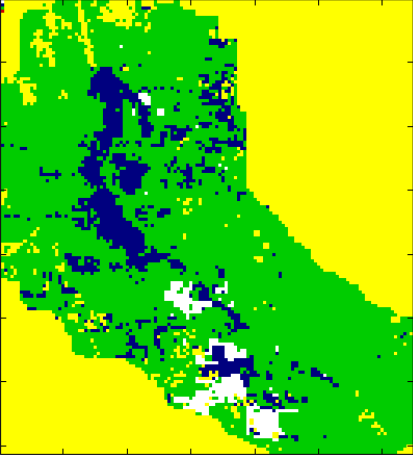}}
    \caption{An illustrative example showing limitations of traditional supervised classification methods for Er Rosieres Reservoir, Sudan on Jun 18th, 2011. (a) False Color Composite image created using bands 7, 5 and 4 from MODIS 8 day product (MOD09GA). (b) Classification map obtained using a standard supervised classification approach}%
    \label{fig:noise}%
\end{figure}

\textbf{Related work on surface water mapping:} Supervised classification of remote sensing images has been widely used for monitoring surface water dynamics and other land cover changes on earth. Specifically, we are considering the setting in which individual snapshots of the region are independently classified with pixels belonging to either land or water class. These individual classification maps are then used to report application specific queries such as extent of lakes, extent variation over time etc. 
Apart from the aforementioned limitation of spatial (MODIS) and temporal (LANDSAT) resolution as described earlier, accuracy of these classification maps are also limited due to factors such as noise and outliers, large amounts of missing data (due to clouds and sensor failures), lack of representative training data, and inadequacy of most classification models to handle the high spatial and temporal heterogeneity at a global scale. 
For example, Figure \ref{fig:noise} shows a classification map (with high degree of inaccuracies) obtained using a standard supervised classification approach described in \cite{khandelwal2017approach}.

\textbf{Our Contributions:} In this paper, we propose to use robust physical principles governing surface water dynamics to transfer information across these noisy classification maps at different scales to produce maps that are not only accurate but also produced at both high spatial and temporal resolution (30m maps at daily scale for the case of MODIS and LANDSAT). Specifically, we propose to use the inherent ordering among instances due to the elevation structure (bathymetry) to create these high quality maps. The effectiveness of this elevation ordering constraint in improving the accuracy of individual classification maps created using any given sensor has been shown in our previous work \cite{khandelwal2017approach,khandelwal2015post} in the context of accurately measuring the surface extent of large reservoirs on a global scale. The key idea is that due to this elevation ordering constraint, if a location is filled with water then due to gravity all the locations in the basin that have lower elevation than the given location should also be filled with water, which enables detection of spurious classification labels that do not adhere to this constraint.  



Our proposed framework, \textbf{ORBIT} (Ordering Based Information Transfer) used this elevation ordering constraint to transfer information across scales and time to achieve our goal of producing HSR surface extent maps at high temporal resolution. The framework has multiple components. First, we propose \orbi\ (Ordering Based Information Transfer Across Scale) that uses elevation ordering at HSR and daily classification maps (that are noisy) at LSR to produce accurate daily classification maps at HSR. The key idea here is that in order to create maps at HSR, we don't need to find labels for all locations at HSR. For example, if we can identify a HSR pixel to be water, then due to elevation ordering constraint, all the locations that are deeper would also be water. 

As mentioned before, daily classification maps can have large amount of noise due to various factors such as clouds, noise and outliers in remote sensing data which can impact the performance of information transfer from LSR to HSR. To overcome this challenge, We propose \orbt\ (Ordering Based Information Transfer Across Time) that aims to exploit rich temporal context available in daily scale classification maps to improve the accuracy of LSR classification maps which would subsequently lead to more accurate information transfer across scale. The key idea here is that in most real situations, a water body grows and shrinks smoothly (except sudden events such as floods) i.e.  surface  extents  of  nearby  dates  are  likely  to  be  very similar. \orbt\  uses elevation ordering to enforce this temporal consistency in total area values which is more robust than enforcing temporal consistency in labels of individual pixels (as done by most current approaches). 

while the focus on this paper is on mapping surface water extents, the ideas presented here are applications in other situations as well where there exists an inherent ordering among instances.

The remainder of the paper is as follows: Section \ref{sec:relwork} provides summary of the related work, Section \ref{sec:background} provides background on elevation ordering based label correction process proposed in \cite{khandelwal2015post}. In Section \ref{sec:itas} and \ref{sec:itat}, we describe the \orbi\ and \orbt\ approach. Section \ref{sec:orbit_proof} provides insights into conditions that are required for effective information transfer across scale in \orbi\ approach. Section \ref{sec:results} provides results and discussion. Finally, Section \ref{sec:future} provides the summary and some promising future research directions.

\section{Related Work}
\label{sec:relwork}
\subsection{Information Transfer Across Time}\hfill

Various post classification label refinement techniques have been proposed that aim to exploit the spatial and temporal context (spatio-temporal auto-correlation) in class labels to detect inconsistent labels and subsequently correct them. A vast majority of work has been done in developing spatial-temporal consistency models using Markov Random Fields to remove "salt and pepper" noise in classification maps. These methods tend to perform poorly in our application because the errors in class labels is also spatially and temporally auto-correlated. Other approaches that incorporate process information have also been developed. For example, in \cite{liu2008using}, authors used transition probability matrices to code domain knowledge about compatible ecological changes in their MAP-MRF model. However, transition probability matrix cannot be assumed or estimated in our application because growing and shrinking of a water body depends largely on external factors such as precipitation, evaporation, in flow, out flow, and ground seepage, etc. that can not be modelled easily.






\subsection{Information Transfer Across Scale}\hfill

Various data fusion techniques have been proposed in remote sensing literature that aim to improve the temporal resolution of fine spatial resolution data by combining information from sensors with differing spatial and temporal characteristics. The most widely used approach is to learn a mapping between raw feature values of two scales and use the mapping to create synthetic HSR image from the corresponding LSR image\cite{gao2006blending,zhu2010enhanced,hilker2009generation}. However these methods have several limitations. First, data from each scale should be precisely calibrated and spectrally normalized to common wavebands which limits their use to datasets with similar multispectral bands. Second, these models learn relationship between pixels at two scales using pair of images at the same date and use it in nearby dates to create synthetic high resolution images. Due to various atmospheric disturbance, a single image pair might not have enough information to learn a robust relationship. Furthermore, these methods cannot handle land cover changes that are not captured in the image pair used to learn the mapping. Finally, these methods are not applicable in situations where there are not overlapping timesteps (e.g. sensors with few days of offset in their revisit cycles). 

A second set of data fusion algorithms is based on unmixing techniques\cite{zurita2008unmixing,amoros2013multitemporal,zhu2016flexible}. Spectral unmixing methods use linear spectral mixture model to extract end members and their proportions on a sub-pixel scale. In unmixing based data fusion techniques, the number of endmembers and proportions is obtained from the high-resolution data set(usually by clustering or classifying high resolution image), and the spectral signature of the endmembers is unmixed from the medium-resolution data set. The issue with these algorithms is that they assume end member proportions do not change between concurrent image pairs and it requires a priori knowledge about the end-members. 


\section{Background}
\label{sec:background}
In \cite{khandelwal2015post, khandelwal2017approach}, we have shown that ordering constraint among instances due to the elevation structure (bathymetry) of a water body provides a very robust physical constraint that can be used to improve the accuracy of the classification maps. The key idea is the following - if a location is filled with water then by laws of physics all the locations in the basin that have lower elevation should also be filled with water. Thus, if we have elevation information then we can detect inconsistent class labels that do not adhere to this physical constraint.   

\begin{figure}[!t]
\centering
\includegraphics[width=0.49\textwidth]{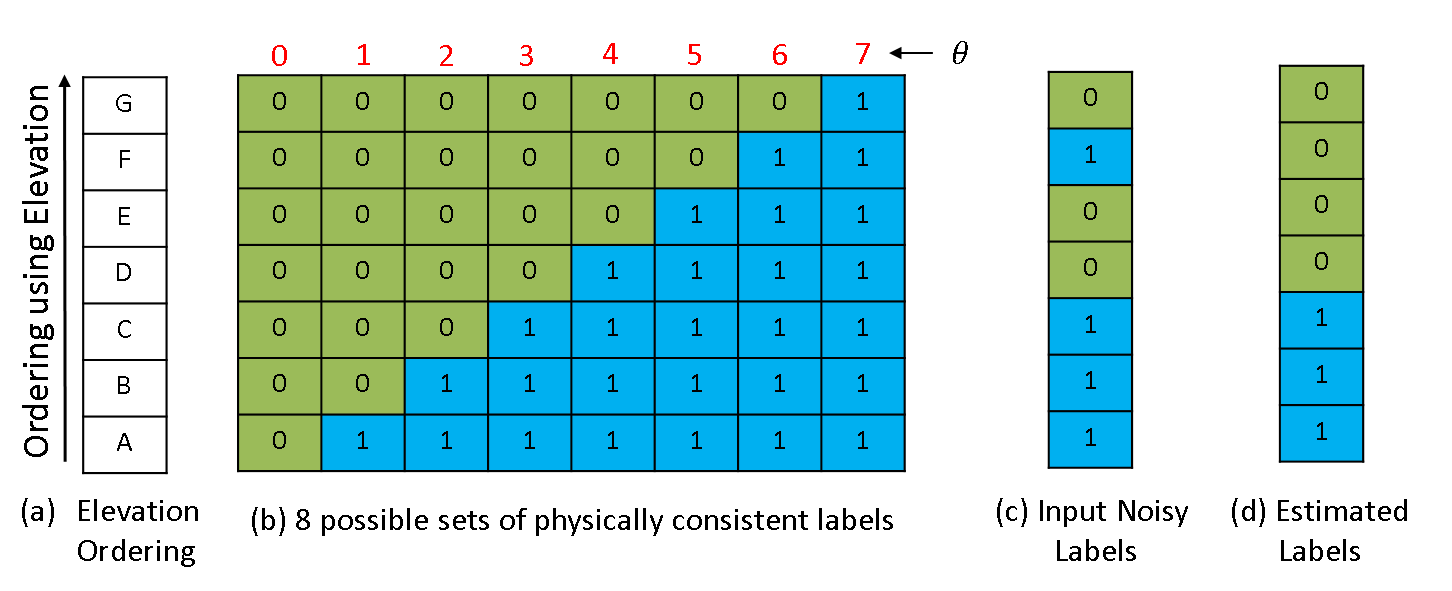}
\caption{An illustrative example showing elevation ordering based label correction process}
\label{fig:waterlevels}
\end{figure}

Here, we describe the methodology proposed in \cite{khandelwal2015post} that uses this elevation ordering constraint to improve the accuracy of labels. Given elevation ordering ($\pi$) and the set of potentially erroneous labels at any given time step $t$, the aim is to estimate correct labels that are physically consistent with the elevation ordering. For a given elevation ordering of N instances, there are only N + 1 possible sets of labels that are physically consistent. For example, figure \ref{fig:waterlevels} (b) shows 8 possible sets of physically consistent labels for 7 locations shown in \ref{fig:waterlevels} (a). Each physically consistent set of labels can also be represented by the number of water pixels ($\theta$) in the set. For example, if we know $\theta$ to be $k$, then by definition it will be the deepest $k$ locations. In the absence of any external information about these labels, the methodology proposed in \cite{khandelwal2015post} adopts the maximum likelihood estimation approach by making an assumption that majority of the input labels are correct and hence selects the set of physically consistent labels that matches the most with input erroneous labels. For example, Figure \ref{fig:waterlevels} (c) shows the erroneous input labels and \ref{fig:waterlevels} (d) shows the selected set that matches the most with input labels. In this illustrative example, location F is detected as erroneous and its label is changed from water to land.

Note that good quality bathymetry information is not explicitly available for most water bodies in the world. To overcome this challenge, the methodology proposed in \cite{khandelwal2015post} estimates this inherent ordering from multi-temporal noisy classification maps using an Expectation-Maximization framework.

\begin{figure}[!t]%
    \centering
    \subfloat[Ground Truth]{\includegraphics[width=0.23\textwidth]{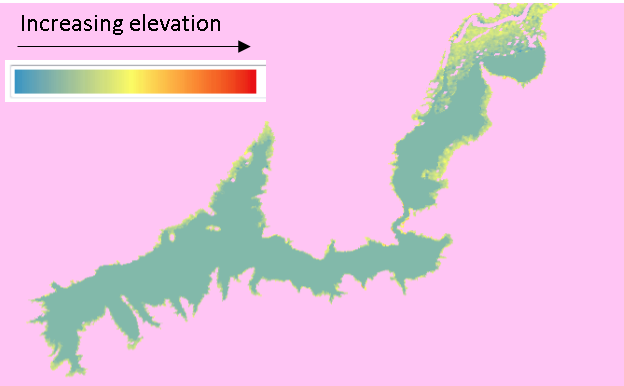}}
    \hfill
    \subfloat[Ground Truth]{\includegraphics[width=0.23\textwidth]{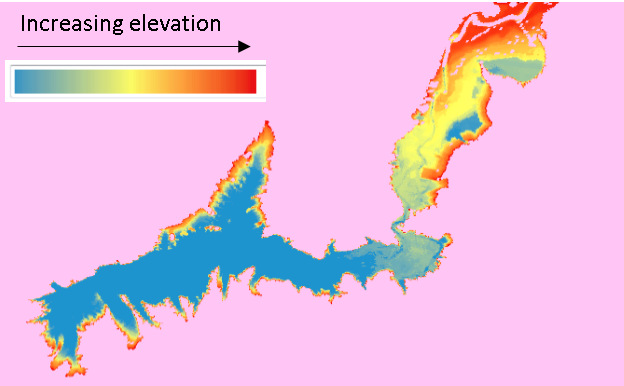}}
    \caption{Comparison of relative elevation orderings at 30m spatial resolution for Kajakai Reservoir, Afghanistan. (a) Relative elevation ordering from SRTM's Digital Elevation data. (b) Relative elevation ordering learned using monthly scale classification maps available from JRC}%
    \label{fig:ele_comp}%
\end{figure}

For example, Figure \ref{fig:ele_comp} (a) shows the relative elevation ordering obtained from a widely used dataset whereas  Figure \ref{fig:ele_comp} (b) shows the relative elevation ordering learned from the data (30 years history (1984 - 2015) of class labels available at monthly scale from JRC product \cite{pekel2016high}) using the methodology proposed in \cite{khandelwal2015post}. As we can see, the elevation ordering learned from the data has much more information than the available elevation ordering. 
Once the ordering is learned, it can be used to correct each classification map individually as described before. In this paper, we will refer to this algorithm as \orbc\ (Ordering Based label CORrection). \orbc\ is a general algorithm that can work with multi-temporal classification maps at any resolution to estimate relative elevation ordering of locations and then use it to produce physically consistent labels. Both \orbi\ and \orbt\ make use of \orbc.


\section{ORBIT-S: Ordering Based Information Transfer across Scale}
\label{sec:itas}
Here, we describe the \orbi\  approach. The key idea behind this approach is to use high level time-independent relationships between instances (elevation ordering in our application) to transfer information across scales. The global nature of elevation ordering constraint enables a more effective transfer of information compared to using local relationship between instances. Specifically, if we can confidently estimate the label of an instance at HSR, then due to the ordering constraint the labels of other instances can be estimated as well.
Furthermore, elevation ordering is an inherent property of locations of the basin and hence does not change with time in most cases and thus enables information transfer even when two dataset are from different duration. 

\subsection{Problem Formulation}\hfill

Given noisy binary classification maps at HSR and low temporal resolution ($\mathbf{H_{i}}$) and noisy binary classification maps at LSR and high temporal resolution ($\mathbf{L_{i}}$) as input, our goal is to estimate accurate classification maps at HSR and high temporal resolution ($\mathbf{\hat{H}_{o}}$). 

\begin{figure}[!t]
\centering
\includegraphics[width=0.45\textwidth, height=4cm]{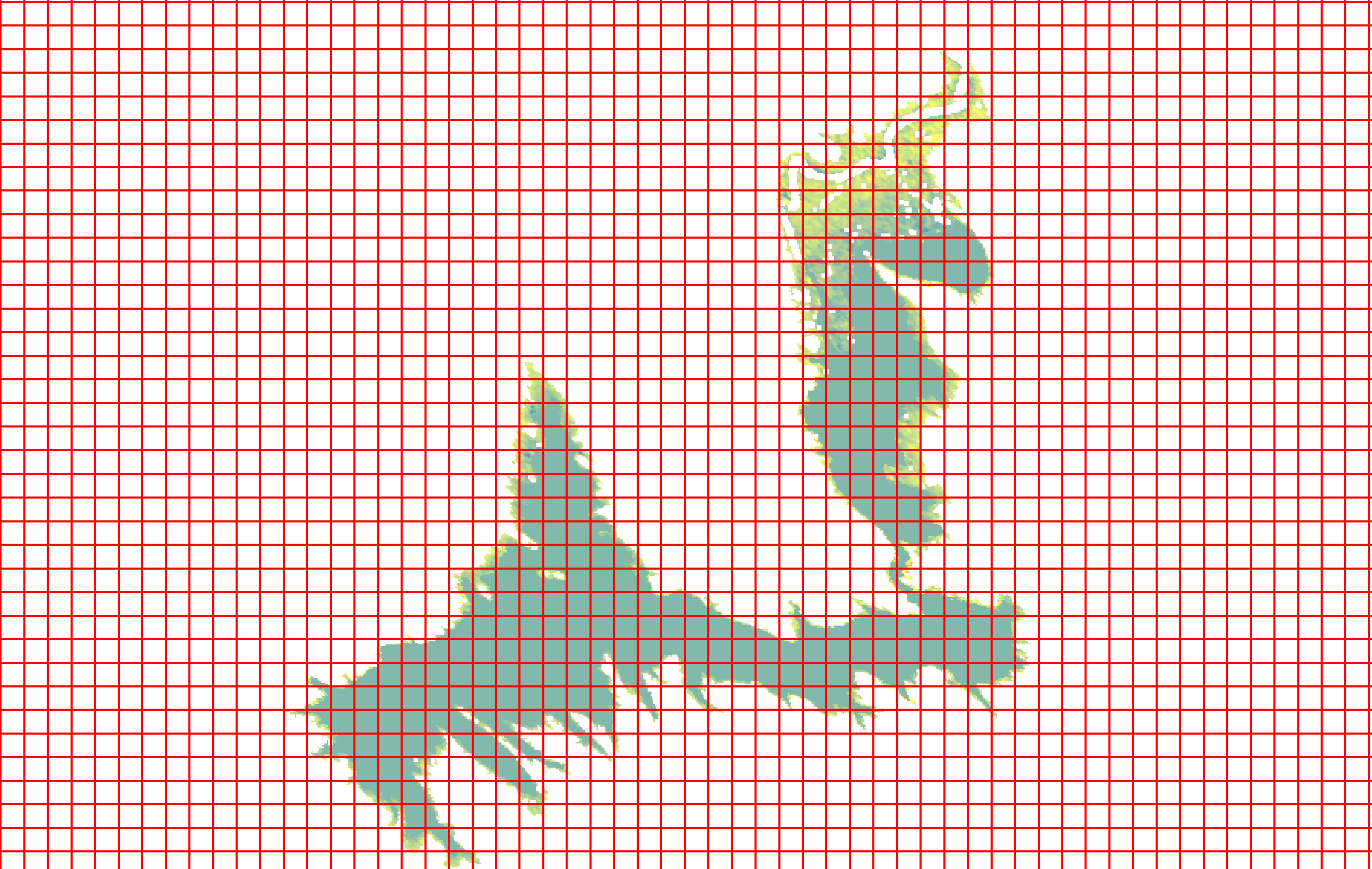}
\caption{Relative elevation ordering for KajaKai Reservoir, Afghanistan from SRTM's Digital Elevation data at 30m spatial resolution. Cells shown in red color corresponding LSR pixels at 600m spatial resolution}
\label{fig:egrid}
\end{figure}

Since, information is available at multiple scales, we define the set of instances at HSR as $P_{h}$ and the set of instances at LSR as $P_{l}$. The mapping grid $G$ defines the mapping between a LSR instance and its corresponding instances at HSR. Specifically, $G_{j}$ represents the set of instances at HSR that belong to the LSR instance $P_{l}^j$. To reduce notations, we assume that each LSR instance has same number of HSR instances within it (defined as $gr$ for grid ratio). For example, Figure \ref{fig:egrid} shows a square grid of 600m spatial resolution overlaid on top of the elevation ordering for Kajakai Reservoir at 30 m spatial resolution.

In order to develop a method to estimate $\mathbf{\hat{H}_{o}}$ from $\mathbf{L_{i}}$, we need to establish the relationship between the label of a LSR instance and the labels of its corresponding HSR instances. In this approach, we make following assumptions:

\textbf{Assumption A1:} The label of a LSR instance is generated by aggregating the labels of its corresponding HSR instances. Specifically, we assume that a LSR instance will be labelled as water if it has more than a certain number of HSR instances (cut-off threshold $wth$) within it labelled as water. 

\textbf{Assumption A2:} In the most general setting, each LSR instance can have a different threshold that can also vary with time due to varying spectral properties of the land cover types enclosed within the LSR instance. Since, the focus of this paper is to demonstrate the utility of using elevation constraint, we make an assumption there exists a single cut-off threshold across all pixels and all timesteps. In the results section, we show that this assumption is very reasonable and in most real situations a global threshold of $gr/2$ performs well. Furthermore, in Section \ref{sec:future} we provide some promising future directions to relax this assumption.    

The \orbi\ approach has five main steps: 1) Estimate elevation ordering at HSR. 2) Estimate elevation ordering at LSR. 3) Estimate accurate and physically consistent classification maps at LSR. 4) Use elevation ordering at HSR from Step 1 and good quality classification maps at LSR from Step 3 to estimate confident labels at HSR and finally 5) Estimate remaining labels at HSR using elevation constraint. Next, we describe these steps in detail.

\subsection{Method}\hfill

Next, we describe the main steps of the \orbi\ approach. 

\textbf{Step 1. Estimate Elevation ordering at high spatial resolution ($\hat{\pi}_{h}$)}\\
In this step, $\mathbf{H_{i}}$ is used to learn elevation ordering using \orbc\ approach. Note that if a high quality elevation structure is available from any external source \orbi\ can directly use it in next steps.

\textbf{Step 2. Estimate Elevation ordering at low spatial resolution ($\hat{\pi}_{l}$)}\\
One way to estimate $\hat{\pi}_{l}$ would be to use \orbc\ with $L_i$ similar to Step 1. However, we estimate $\hat{\pi}_{l}$ using $\hat{\pi}_{h}$ and $\mathbf{L_{i}}$ as it allows us to 1) estimate the threshold $wth$ together with $\hat{\pi}_{l}$ and 2) ensure that ordering learned at LSR is coherent with ordering at HSR. 

As mentioned before, each LSR pixel contains $gr$ number of HSR instances where each of the $gr$ instances have a ranking from $\pi_h$. Using $\pi_h$, LSR instances can be ranked in number of ways. For example, LSR instances can be ranked based on the lowest rank with each LSR instance. Similarly, they can be ranked using the highest rank within each LSR instance. Here, we proposed to generate possible LSR orderings on the basis on assumption A1. Specifically, we define a LSR ordering $\pi_{l}^{wth}$ as the ordering obtained by using HSR instances with local rank $wth$ within each LSR instance. Thus, there can be $gr$ possible LSR orderings that can be generated from $\pi_h$. 

In the absence of any external information about the correct labels, we select that LSR ordering that leads to the least amount of corrections in $\mathbf{L_{i}}$. This would also automatically provide the estimate of $wth$ that will be used in next steps of the algorithm. Specifically, $wth$ value corresponding to the selected LSR ordering is chosen as the cut-off threshold.

\textbf{Step3: Estimate accurate and physically consistent classification maps at LSR ($\mathbf{L_{o}}$)}\\
Once the LSR ordering is estimated in the previous step, we can use it to correct each classification map ($L_{i}^{t}$) individually to obtain physically consistent classification maps at LSR ($L_{o}^{t}$) using \orbc. This step plays an important role because if the information in these LSR maps is of bad quality then it will get propagated in the estimated HSR maps as well.

\textbf{Step 4. Estimate confident labels in $\mathbf{\hat{H_o}}$}\\
According to assumption A1, a LSR instance can be labelled as water only if it has at least $wth$ HSR water instances. This implies that if we are given a LSR instance labelled as water, then at least $wth$ HSR instances within it should be water. By definition, these $wth$ instances will be filled according to their elevation rank (deeper to shallower). Similarly, if a LSR instance is labelled as land then at least $gr - wth$ HSR instances within it should be land. Using this knowledge, we can confidently estimate physically consistent labels for some of the HSR instances within each LSR instance. 

\begin{figure}[!t]
    \centering
    \subfloat[]{\includegraphics[width=0.3\textwidth]{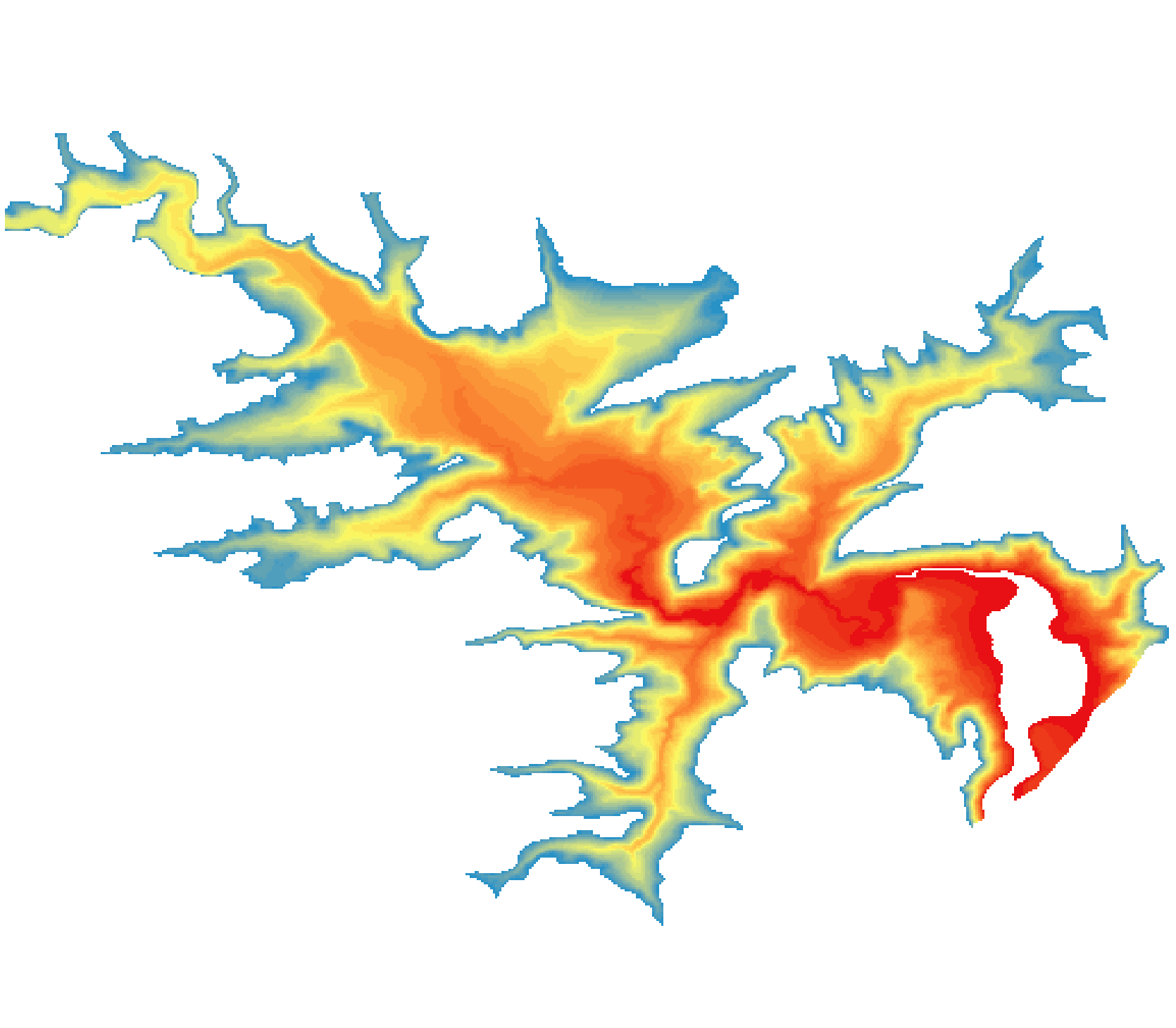}}%
    \qquad
    \subfloat[]{{\includegraphics[width=0.2\textwidth]{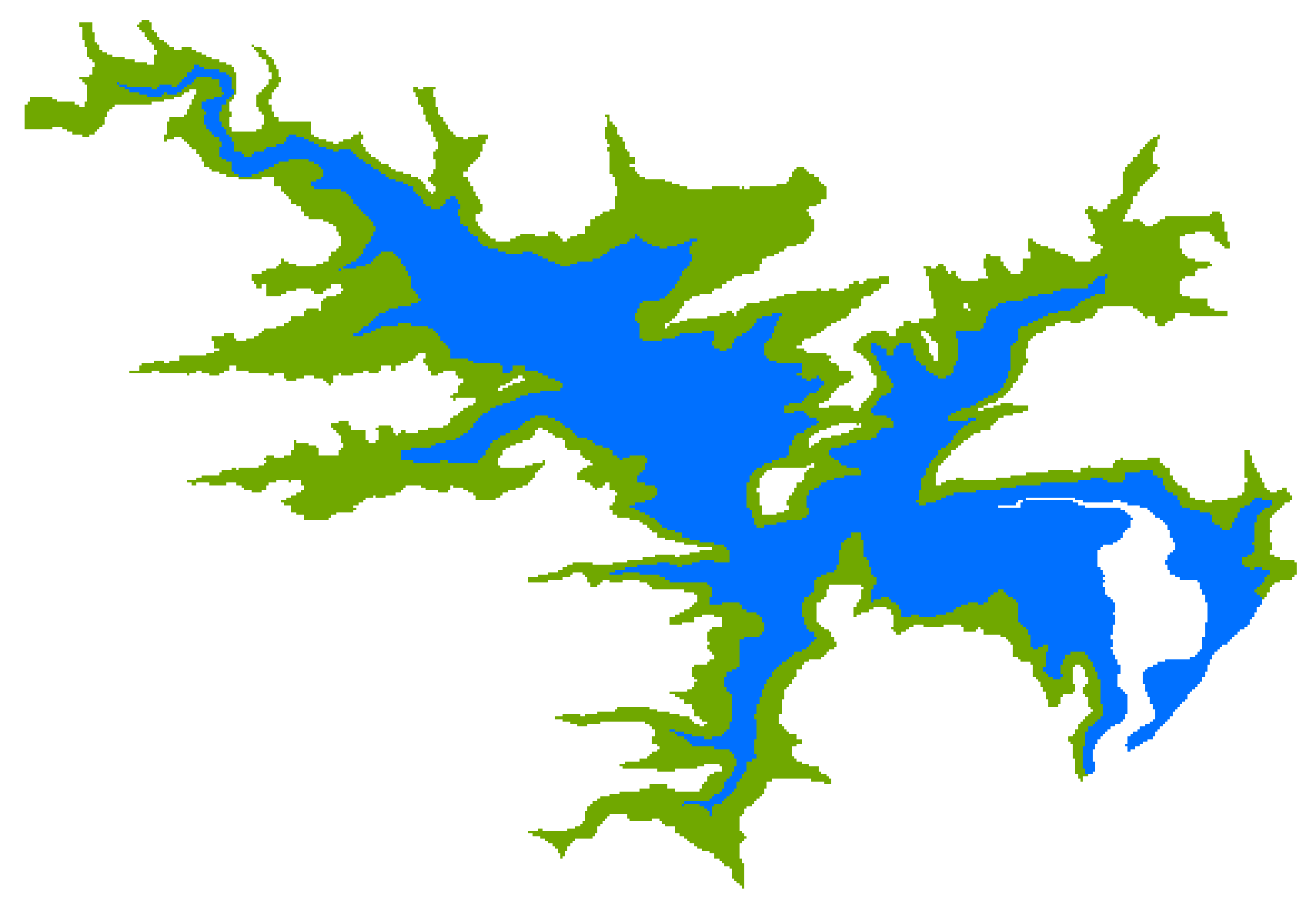} }}%
    \subfloat[]{{\includegraphics[width=0.2\textwidth]{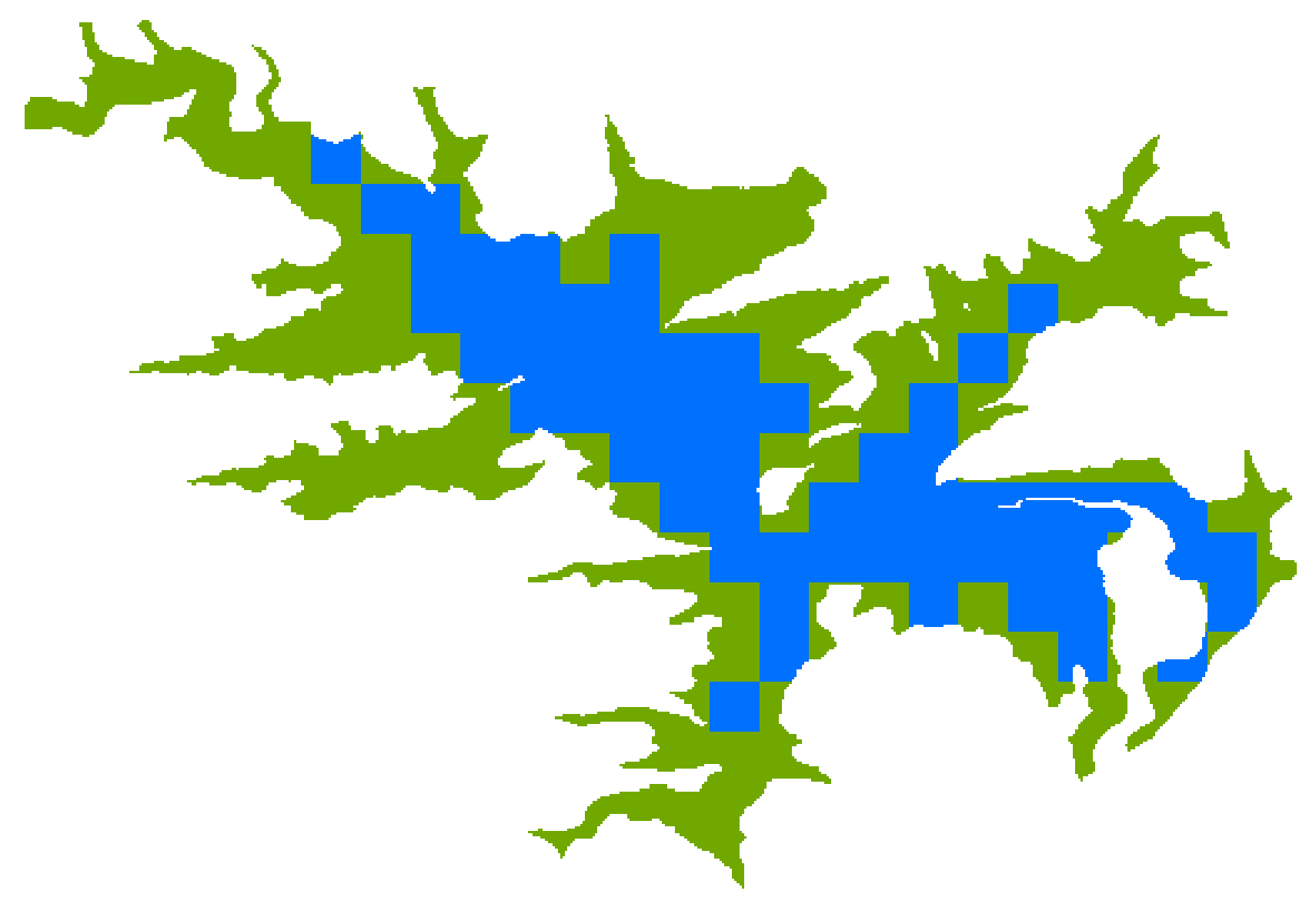} }}%
    \qquad
    \subfloat[]{{\includegraphics[width=0.2\textwidth]{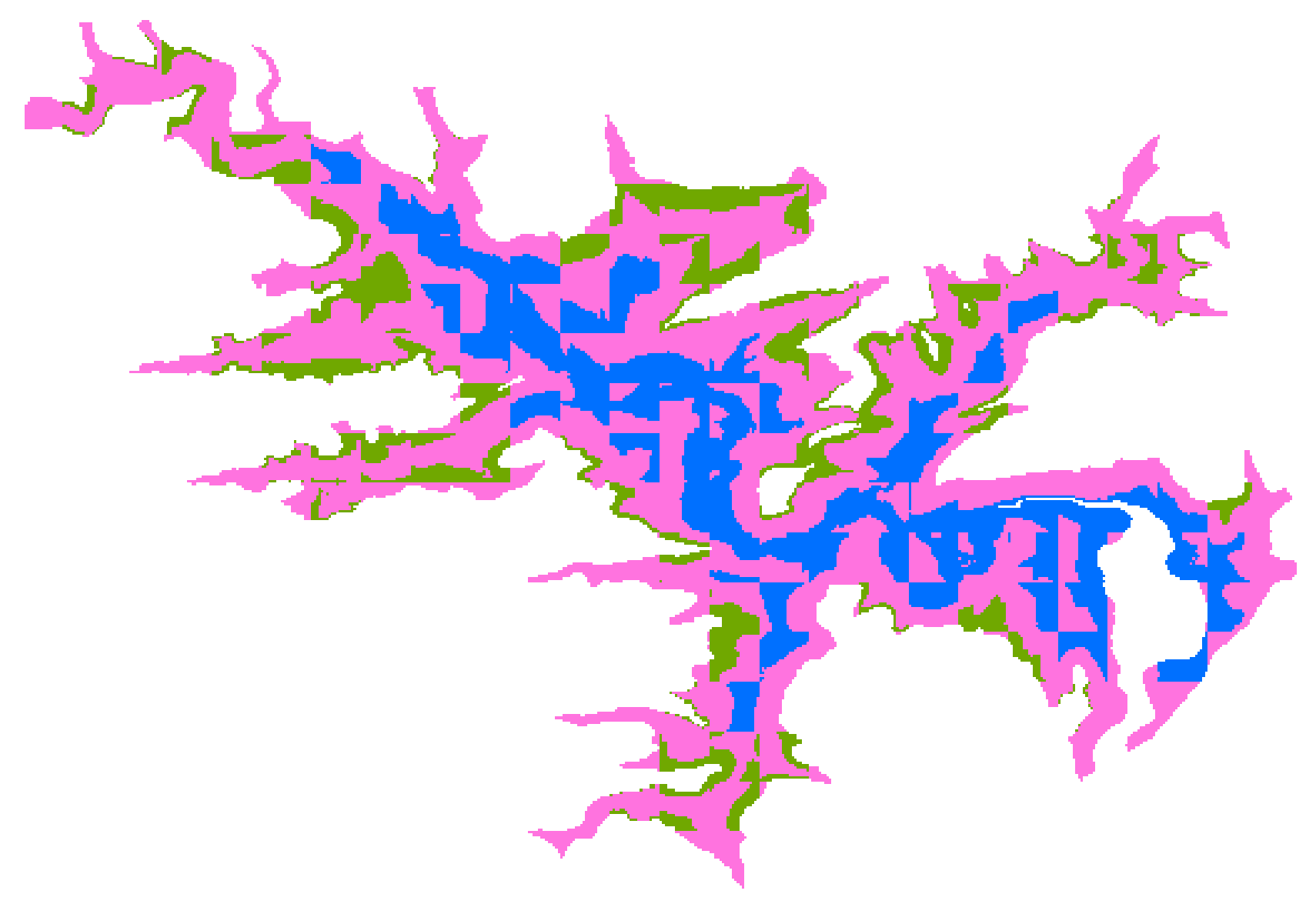} }}%
    \subfloat[]{{\includegraphics[width=0.2\textwidth]{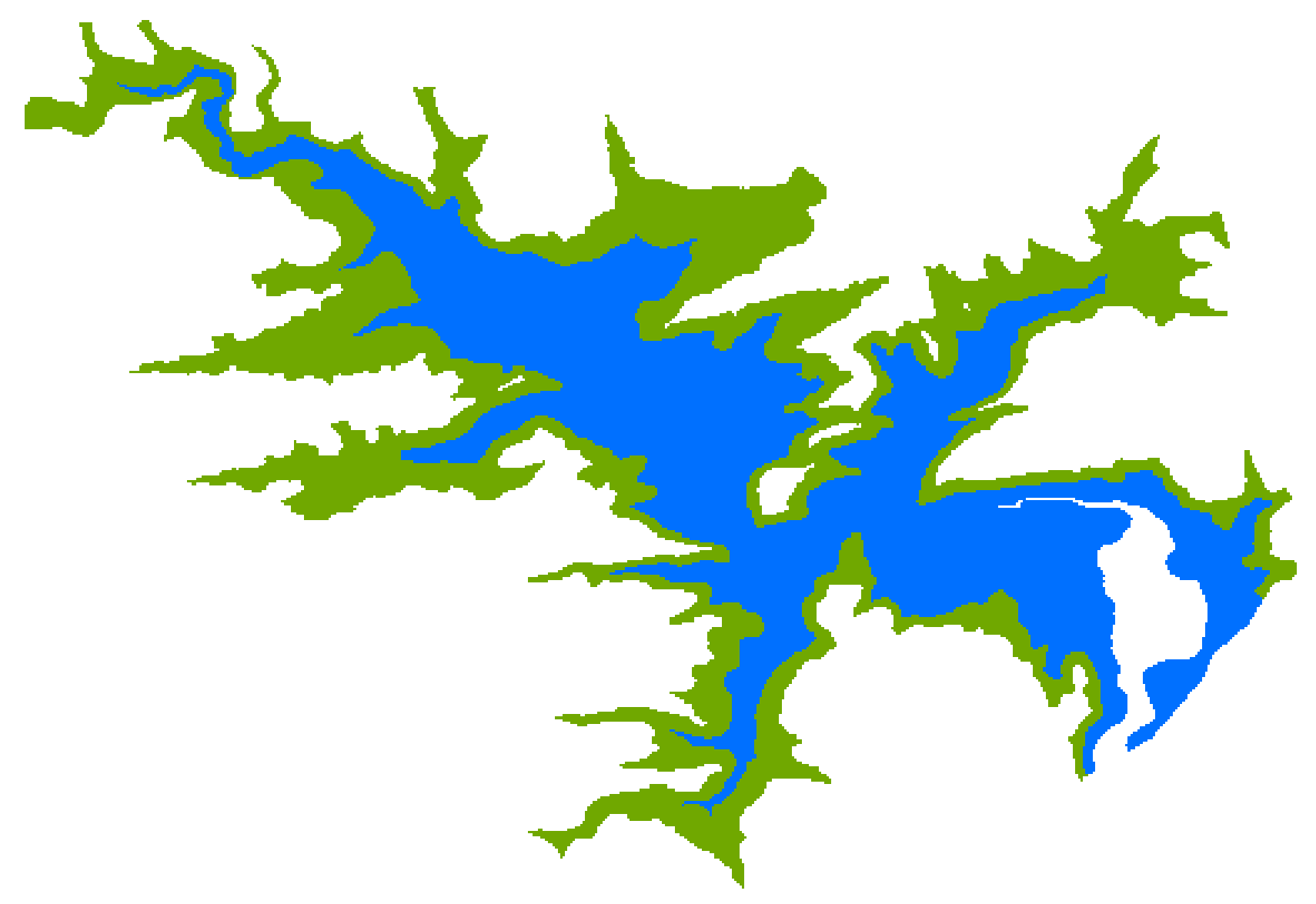} }}%
    \caption{An example showing improvement in spatial resolution of surface extent maps for Red Fleet Reservoir, Afghanistan. (a) High Quality bathymetry at 10m spatial resolution. Synthetically create HSR map using bathymetry. (c) Perfect LSR map at 200m. (d) Estimated HSR map at 10m after Step 4 (e) Estimated HSR map at 10m after Step 5. In this example, \orbi\ was able to  estimate the HSR map perfectly }%
    \label{fig:orbi_example}%
\end{figure}

\textbf{Step 5. Estimate remaining labels in $\mathbf{\hat{H_o}}$}\\
After Step4, there will be a lot of unknown labels in $\mathbf{\hat{H_o}}$. For example, if $wth$ is $gr/2$, then half of the labels in $\hat{H_o}$ would be unknown after Step 4. In this final step, we use $\hat{\pi}_h$ to estimate the labels of remaining instances. Specifically, we first find the shallowest HSR pixel that is labelled as water ($Pivot_w$) and label all the instances deeper than it as water as well due to the physical constraint. Using the same rationale, we find the deepest HSR instance labelled as land ($Pivot_l$) and label all instances shallower than it as land as well. This step significantly reduce the number of unknown labels. Finally, instances that are between pivots $Pivot_w$ and $Pivot_l$ remain unlabeled. Note that the elevation of $Pivot_l$ will always be higher that the elevation of $Pivot_w$ because Step 4 ensures that only physically consistent labels are estimated. Ideally, the gap between $Pivot_l$ and $Pivot_w$ should be as small as possible. In section \ref{sec:orbit_proof} we provide some insights about scenarios where this gap would be small. Figure \ref{fig:orbi_example} shows an illustrative example showing different steps of the \orbi\ approach for Red Fleet Reservoir in USA.


\section{\orbt\: Ordering Based Information Transfer across Time}
\label{sec:itat}
Step 3 of the \orbi\ approach uses \orbc\ to correct each LSR classification map ($L_{i}^{t}$) individually to obtain physically consistent classification map at LSR ($L_{o}^{t}$). Since, MODIS based classification maps at daily scale can have lots of errors and missing data due to clouds and other atmospheric disturbances, correcting each time step independently can lead to abrupt spurious changes in surface extents. As mentioned before, the accuracy of LSR maps play an important role in the performance of the \orbi\ approach because errors in label at LSR would propagate to HSR. Hence, it is important to ensure the accuracy of the estimated LSR maps.

\begin{figure}[!t]
\centering
\includegraphics[width=0.45\textwidth]{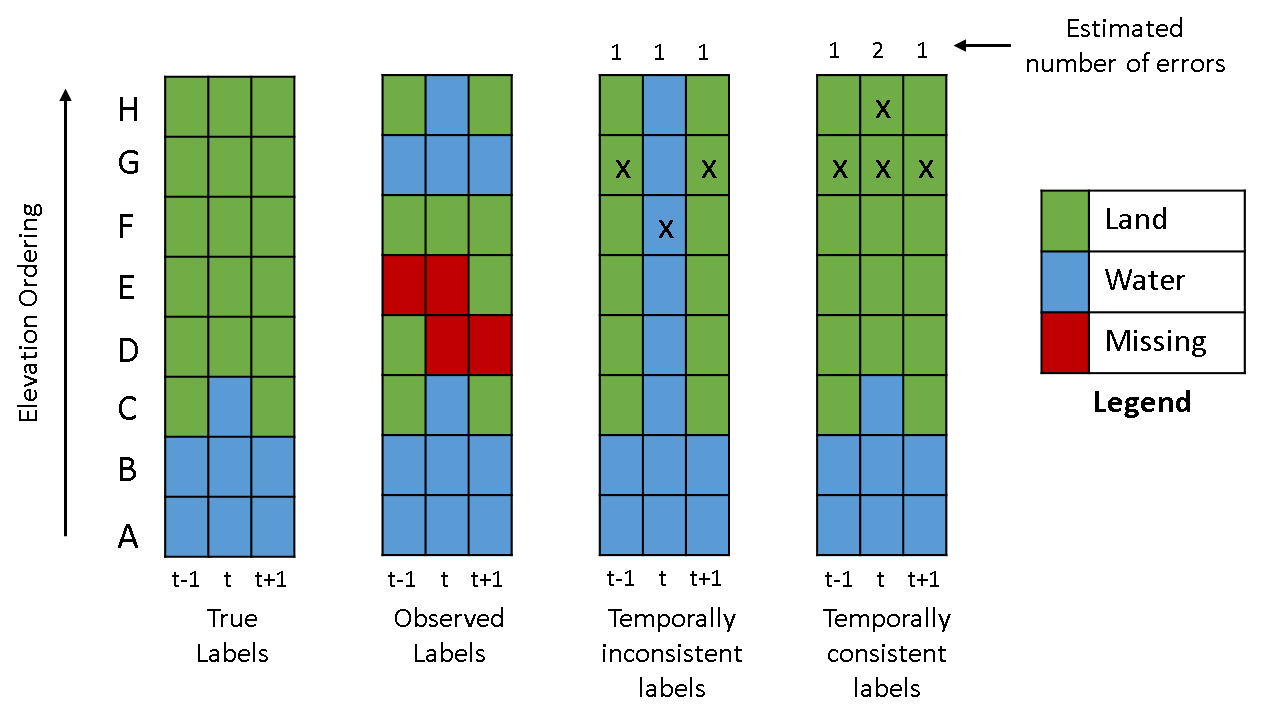}
\caption{An illustrative example showing the utility of incorporating temporal context in elevation based label correction process.}
\label{fig:DPFigure2}
\end{figure}

In most situations, a water body grows and shrinks smoothly (except sudden events such as floods) i.e. surface extents of nearby dates are likely to be very similar. Hence, incorporating temporal context in the label correction process can lead to better performance. Consider the illustrative example shown in figure \ref{fig:DPFigure2}. Figure \ref{fig:DPFigure2} (c) shows corrected labels when each time step is corrected independently which leads to abrupt increase in number of water locations in time step $t$. Figure \ref{fig:DPFigure2} (d) shows labels that are more temporally consistency which can be obtained by increasing the estimated number of errors by just 1.

Current state-of-the-art methods mainly enforce the temporal consistency either for each pixel individually (e.g. majority filters in time) or use a given pixel's temporal and spatial neighborhood to obtain temporal consistent labels. As shown in \cite{khandelwal2015post}, these methods perform poorly when noise and missing data is also spatially and temporally auto-correlated which is very common in our application. For example, in Figure \ref{fig:DPFigure2}, the errors in locations $G$ are likely to remain uncorrected by existing methods. Moreover, existing methods tend to remove real changes in labels as well as they enforce labels in nearby time steps to be similar. 

In this paper, we proposed to use elevation ordering to enforce temporal consistency. Specifically, we propose to use elevation ordering to enforce temporal consistency in total area values instead of consistency in labels of individual pixels. Temporal consistency in total area (surface extent) is more direct constraint and it also allows real change to happen more easily than existing methods. Enforcing consistency in total area is very natural in our approach because elevation based label correction method estimates water level each time step already.

\subsection{Problem Formulation}\hfill

Given, an elevation ordering ($\pi$) and a set of noisy classification maps $\mathbf{C_i}$, our goal is to estimated physically consistent maps $\mathbf{C_o}$. If a water body has $N$ locations then there are $N+1$ possible water levels for each time step as described in section \ref{sec:background}. Further, if the water body has been observed for $T$ time steps, then there are exponential ($N^T$) possible configurations of water levels for the given water body. Our goal is to find that configuration of water levels that not only show good consistency with input labels (minimizes number of corrections/mismatches in input labels) but also show good temporal consistency (smoothly changing water levels). 

We define the measure of mismatch between input labels and estimated physically consistent labels as 

\begin{equation}
Cost_{mismatch}(T) = \sum_{t=0}^{T}Err_{\pi}(\theta_{t})
\label{eq:cost_mismatch}
\end{equation}
where, $Err_{\pi}(\theta_{t})$ represents number of mismatches between input labels and physically consistent labels at timestep $t$ when the water level is chosen to be $\theta_{t}$
This is the same cost function used in \orbc. Next, we define the measure for temporal consistency as 
\begin{equation}
Cost_{transition}(T) = \sum_{t=0}^{T-1}|(\hat{\theta_{t}} - \hat{\theta_{t+1}})|
\label{eq:cost_transition}
\end{equation}

$Cost_{transition}$ represents an aggregate measure of abrupt changes in water levels till time $T$ where $|(\hat{\theta_{t}} - \hat{\theta_{t+1}})|$ represents absolute difference in water levels at consecutive time steps $t$ and $t+1$. The above criteria enforces temporal consistency as it favours similar water levels in nearby time steps. 

Both goals (Eqn \ref{eq:cost_mismatch} and Eqn \ref{eq:cost_transition}) conflict with each other. The configuration that leads to best consistency with input labels is obtained when each time step is corrected independently (as explained in section \ref{sec:background}) but would lead to least temporal consistency in water levels. On the other hand, the most temporally consistent configuration is obtained when all the levels are forced to be same (no dynamics in surface extents) but would lead to poor consistency with input labels. Hence, we define the overall objective function as 

\begin{equation}
\argmin_{\hat{\theta_{1 ... T}}} Cost_{total}(T)
\label{eq:obj}
\end{equation}
where, 
\begin{equation}
Cost_{total} = Cost_{mismatch}(T) + \alpha Cost_{transition}(T) 
\label{eq:cost_total}
\end{equation}
and $\alpha \in [0,]$ is the trade-off parameter between two costs. 

As $\alpha$ is increased, the above criteria will favour water levels that are more temporally consistent. Hence, $Cost_{transition}$ decreases as $\alpha$ is increased.  As mentioned before, $Cost_{mismatch}$ is minimum when water levels for each time steps are estimated independently (i.e. $\alpha=0$). Hence, as $\alpha$ is increased, $Cost_{mismatch}$ increases.




\subsection{Method}\hfill

We propose a dynamic programming formulation to optimally solve the above objective for any given value of $\alpha$. The objective function in Eqn \ref{eq:cost_total} can be written in expanded form as 

\begin{equation}
\begin{split}
Cost_{total}(T) & = \sum_{t=0}^{T}Err_{\pi}(\theta_{t}) + \alpha \sum_{t=0}^{T-1}|(\hat{\theta_{t}} - \hat{\theta_{t+1}})|\\
\\
 & = \sum_{t=0}^{T-1}Err_{\pi}(\theta_{t}) + \alpha \sum_{t=0}^{T-2}|(\hat{\theta_{t}} - \hat{\theta_{t+1}})|\\
 & + Err_{\pi}(\theta_{T}) + \alpha |(\hat{\theta_{T-1}} - \hat{\theta_{T}})|\\
\\
 & = Cost_{total}(T-1) + Err_{\pi}(\theta_{T}) \\
 & + \alpha |(\hat{\theta_{T-1}} - \hat{\theta_{T}})|
\end{split}
\label{eq:rec}
\end{equation}




Eqn \ref{eq:rec} shows the recursive form of the objective function. This recursive equation can be minimized using dynamic programming algorithm. The path that leads to the minimum value of the objective function are chosen as the estimated water levels.

Thus, \orbt\ provides provides the ability to effectively model and control the impact of temporal consistency using the parameter $\alpha$. Since, \orbt\ is an alternate approach to \orbc\ for obtaining physically consistent labels, we define a new variation of \orbi\ (henceforth referred to as \orbst) that uses \orbt\ instead of \orbc\ in Step 3 of \orbi.

\section{Theoretical Analysis of \orbi}
\label{sec:orbit_proof}
As mentioned earlier, after the final step of \orbi, there can be two types of issues with the estimated HSR maps - 1) propagation of error from erroneous LSR labels to HSR labels and 2) number of instances with unknown label (gap between two pivots ($Pivot_l - Pivot_{w}$). The first type of error depends on the quality of input LSR labels and hence will reduce as their quality increases. However, number of unknown labels depend on factors that are independent of data such as shape and size of the lake, its elevation structure and difference in the resolutions of the two scales. In this section, we provide insights into the impact of these characteristics on the number of unknown labels produced by \orbi\ approach. Here we assume that LSR labels are perfect. 

The key reason behind the success of Step 5 of \orbi\ approach is that a water body grows and shrinks in layers (contours). Given this nature of the elevation structure, if we can identify a water pixel in any contour, then we can be certain that all the deeper contours will also be filled with water. Similarly if we can identify a land pixel in any contour, then we can be certain that all the shallower contours will have no water. So, if we can identify a water pixel in the last contour filled with water (henceforth referred to as boundary water pixel, $B_{w}$) and similarly a land label in the first empty contour (henceforth referred to as boundary land pixel, $B_{l}$), then there will be no unknown labels in the HSR map.

Here, we provide a probabilistic bound on ability of \orbi\ approach to identify boundary water and land pixels.

\textbf{Identification of a boundary water pixel:} Step 4 of \orbi\ approach estimate labels at HSR using the assumption A1. Specifically, if a LSR instance is labelled as water then all the HSR instances within it with local rank $\leq wth$ are labelled as water. Consider a boundary water pixel at HSR, $B_w$ that belongs to the LSR pixel $P_l^j$. According to assumption A1, $B_w$ will be labelled as water in Step 4 only if 1) $P_l^j$ is labelled as water and 2) $B_w$ has local rank $\leq wth$. However, both these conditions are contradictory because boundary water pixels are the last pixels that get filled with water by definition and hence determine label of the corresponding LSR pixel. For example, if $B_w$ has local rank $\leq wth$ then the total number of water pixels in $P_l^j$ will be less than $wth$ and hence cannot have water label. In fact, both conditions satisfy simultaneously only when $B_w$ itself has local rank $= wth$. In this case, the $P_l^j$ will have \textbf{exactly} $wth$ water pixels and hence will get a water label. To summarize, a boundary water pixel can be identified by \orbi\ approach if the corresponding LSR pixel has \textbf{exactly} $wth$ HSR water pixels in it. (Note that this is not the only condition in which boundary pixels can be detected. In order to do the worst case analysis, here we assumed that each LSR pixel will have only one HSR boundary water pixel. But in reality, a LSR pixel can have more than one HSR boundary water pixels and which implies that there can be boundary water pixels that have local rank $\leq wth$ also belong to a LSR pixel labelled as water.) 

A LSR pixel can have gr + 1 possible number of HSR water pixels in it. ($gr$ is the number of HSR instances in a LSR instance). Thus, the probability of a LSR pixel to have exactly $wth$ HSR water pixels can be defined as  $\frac{1}{(gr+1)}$. As mentioned before, number HSR water pixels in a LSR pixel depends on the local rank of the boundary water pixels. Since, in general the shape of the lake is not related to the imposed mapping grid (Assumption A3), local rank of boundary pixels in a LSR pixel does not determine local rank of boundary pixels in other pixels the probability of detecting a water pixel in the last filled contour can be defined as 

\begin{equation}
    1 - (1- \frac{1}{gr+1})^C
\end{equation}
where, $(1- \frac{1}{gr+1})$ is the probability that a LSR pixel does not have exactly $wth$ HSR water pixels and $C$ is the number of LSR pixels intersecting the last filled contour (.i.e. the perimeter of the lake extent at LSR).

To make this more general, consider the situation where there are no LSR pixels with exactly $wth$ HSR water pixels. In this case, we won't be able to detect the last filled contour. However, if there are LSR pixels with $wth+1$ water pixels then we can identify a HSR water pixel in the second to last filled contour. In general, the probability that the detected water contour will be within k levels deeper than the true water contour can be defined as - 
\begin{equation}
    1 - (1- \frac{k+1}{gr+1})^C
    \label{eq:proof1}
\end{equation}

\textbf{Identification of a boundary land pixel:} Following the above discussion, $B_l$ that belongs to the LSR pixel $P_l^j$ will be detected as land only if $P_l^j$ has \textbf{exactly} $wth-1$ water pixels. In this case, step 4 of \orbi\ approach will label all the remaining HSR pixels ($gr - (wth-1)$) as land which will include $B_w$ by definition. Similar to case for detecting boundary water pixels, the probability that the detected land contour will be within k levels shallower than the true land contour can be defined by \ref{eq:proof1} as well.


Thus, the probability that the unknown labels will be restricted within k contours around the true boundary contours is 

\begin{equation}
    (1 - (1- \frac{k+1}{gr+1})^C) * (1 - (1- \frac{k+1}{gr+1})^C)
    \label{eq:proof2}
\end{equation}

\begin{figure}[!t]
\centering
\includegraphics[width=0.3\textwidth]{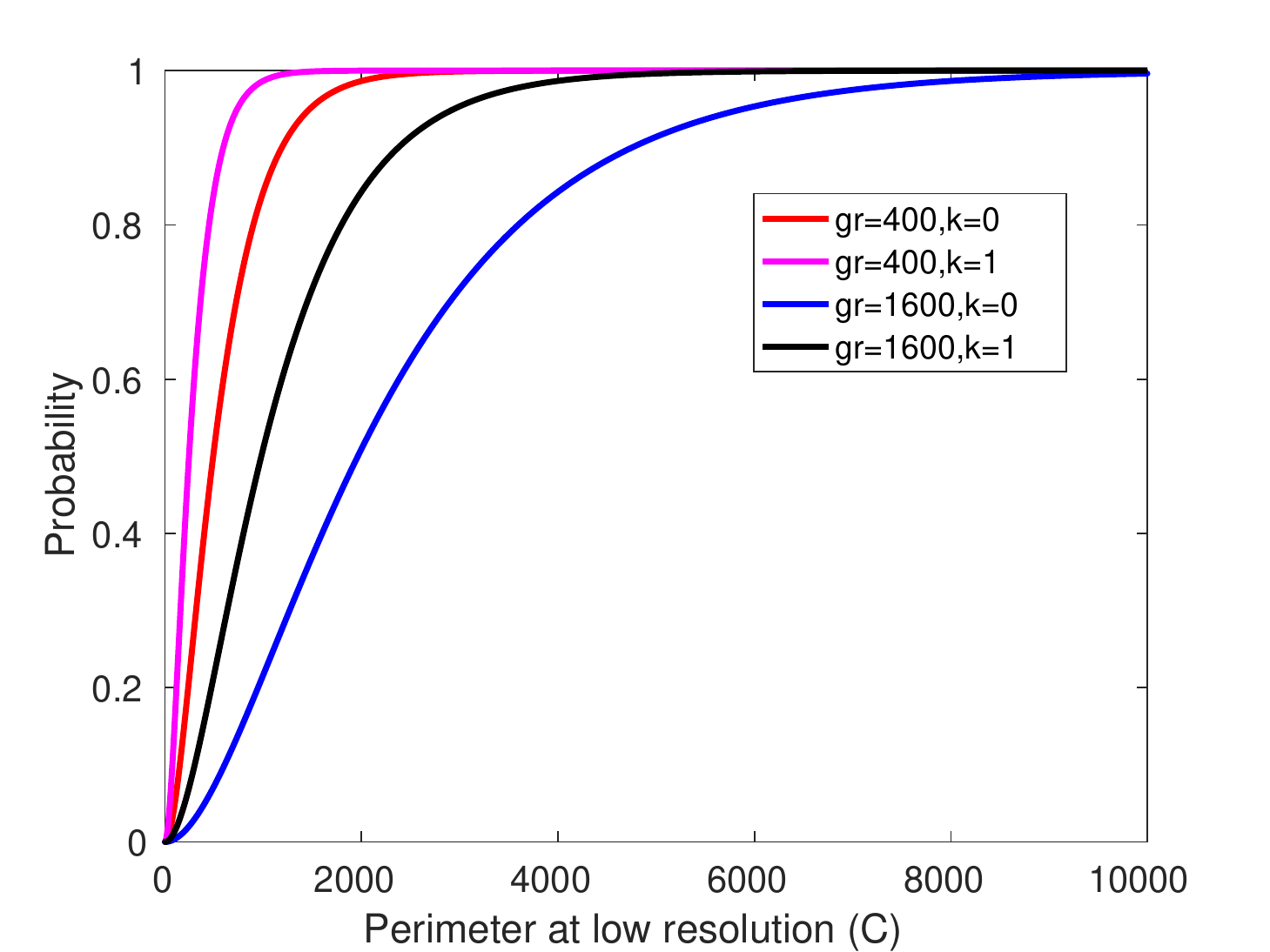}
\caption{Relationship between lake perimeter and number of unknown pixels labeled by \orbi}
\label{fig:PACcurves}
\end{figure}

As we can see, the probability that the unknowns will be localized with k layers increase as the perimeter at low resolution increases. On the other hand, the probability decreases as ratio between scales is increased. Note that the probability does not depend on the value of $wth$ due to assumption A3. In the results section, we will provide empirical evidence for the validity of assumption A3. 
Figure \ref{fig:PACcurves} show the relationship between probability of detecting water and land boundary pixels for two different values of $gr$ and $k$. 

It may appear that the conditions required to limit the unknown labels to boundary contours are very specific and may not hold true in real situations. But these conditions hold true in most cases as we have shown empirically on real world lakes in section \ref{sec:results}. 

\section{Results}
\label{sec:results}
In this section, we evaluate the proposed approaches using both synthetic datasets as well as real-world datasets. Note that we have not compared the proposed framework with other baselines because to the best of our knowledge there are no existing methods that can do information transfer across scale using classification maps.  

\subsection{Synthetic Data Experiments} \hfill

\begin{figure}[!t]
\centering
\subfloat[]{{\includegraphics[width=0.49\textwidth]{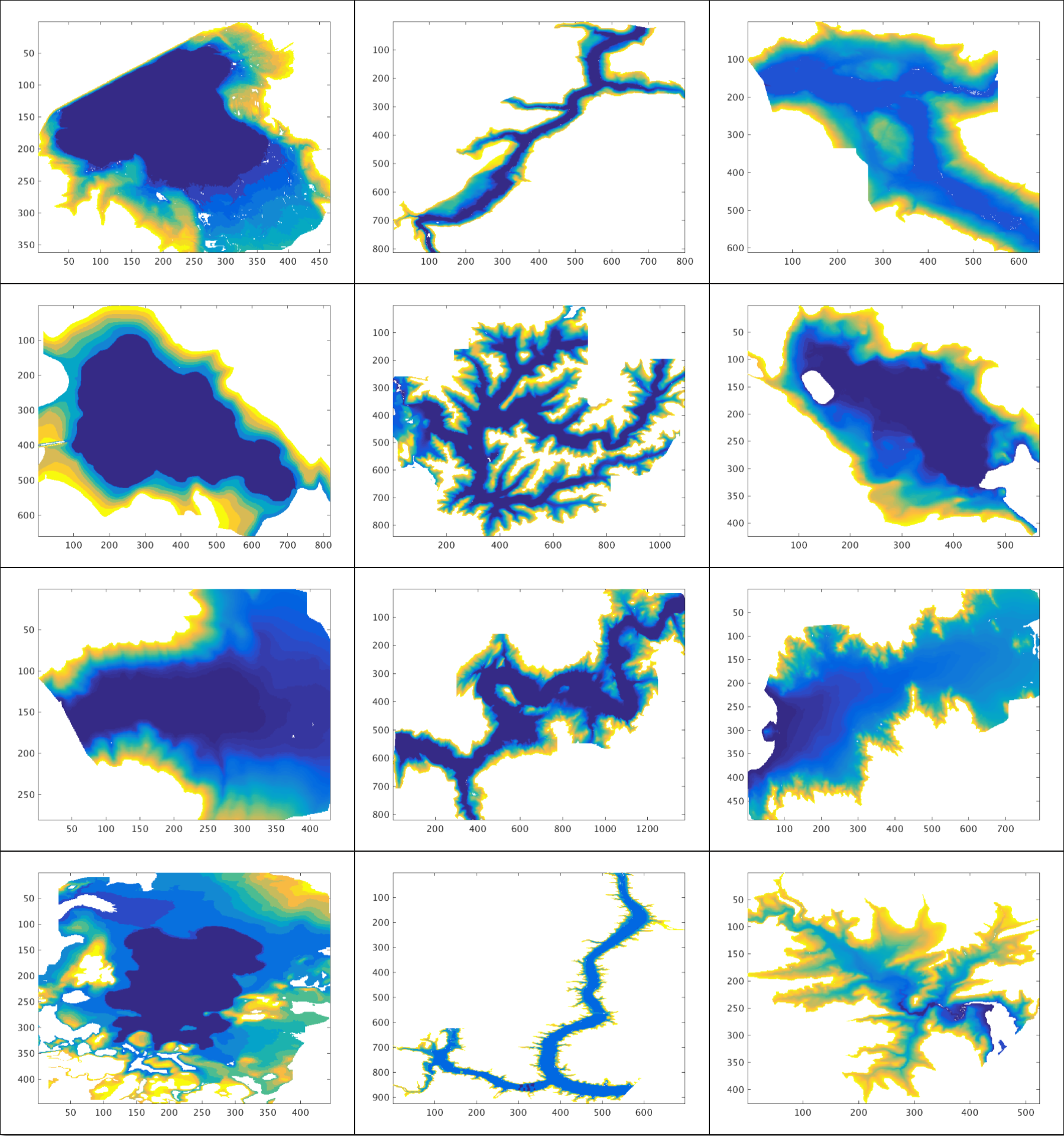}}}%
\qquad
\caption{Elevation structure of a sample of 12 lakes from the 620 lakes dataset used in the paper.}
\label{fig:dem10collage}
\end{figure}

\textbf{Impact of Shape and Size:} In this experiment, we will provide empirical analysis of the bound given in section \ref{sec:orbit_proof}. The goal of this experiment is to analyze the amount of unknown labels produced during the information transfer process. We obtained elevation structure (bathymetry) for 620 lakes of varying shapes and sizes from USGS's Digital Elevation data which is available at 10m spatial resolution for most parts of the USA. Figure \ref{fig:dem10collage} shows a sample of 12 lakes of different shapes and size out of the 620 lakes used in the paper. Note that some of the water bodies have a flat blue region. This is due to the flattening artifact that happens in using radar instruments to map the terrain. Since, the microwave signal used cannot penetrate the water surface, there is no elevation information present under the extent at which lake existed when the data was collected. 

We created multiple surfaces extents ($H_{gt}$) at HSR from these lakes. Specifically, for each lake we created surface extents by filling the lake's bathymetry at different levels. To simulate the LSR scale, we created  synthetic square grids of 200m ($gr$ = 400) and 400m ($gr$ = 1600) spatial resolution and used it to create corresponding accurate and physically consistent LSR surface extent maps ($L_{gt}$) for two different values of $wth$ ($0.5*gr\ and\  0.75*gr$). We then used the elevation ordering and $L_{gt}$ to estimate HSR maps ($H_{est}$). These HSR maps will have unknowns in them but no erroneous labels because $L_{gt}$ have no errors. 

\begin{figure}[!t]
\centering
\subfloat[$gr = 400, wth = 200$]{{\includegraphics[width=0.2\textwidth]{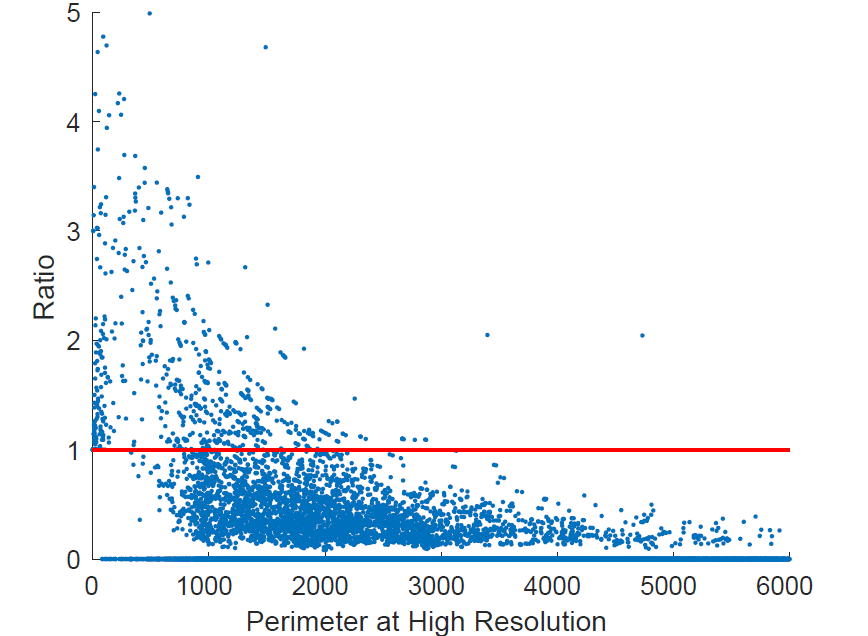}}}%
\subfloat[$gr = 400, wth = 300$]{{\includegraphics[width=0.2\textwidth]{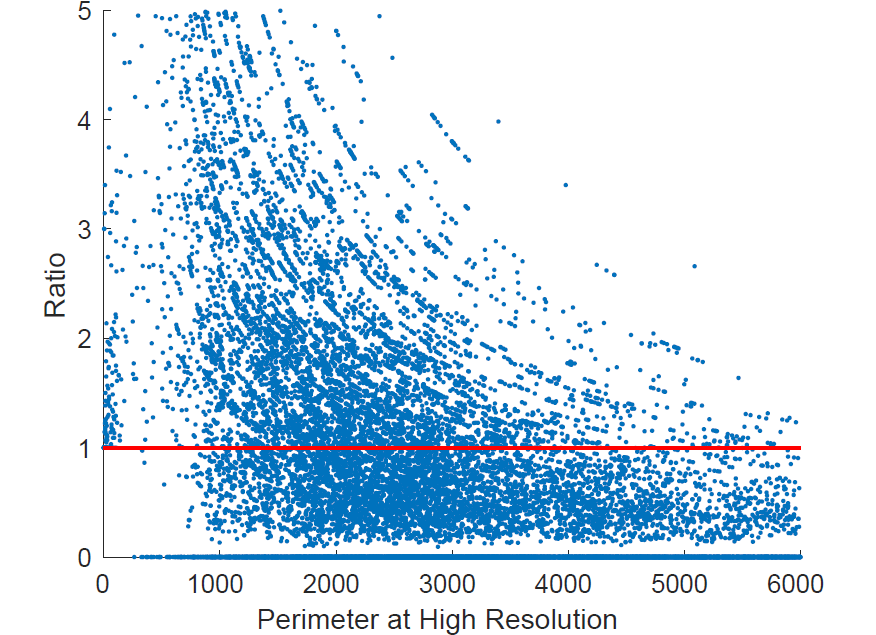} }}%
\qquad
\subfloat[$gr = 1600, wth = 800$]{{\includegraphics[width=0.2\textwidth]{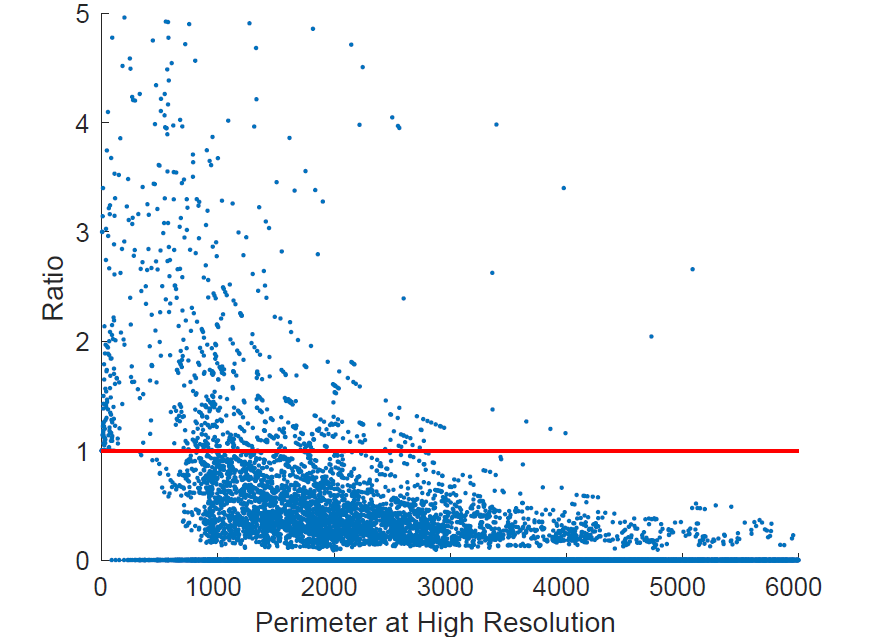} }}%
\subfloat[$gr = 1600, wth = 1200$]{{\includegraphics[width=0.2\textwidth]{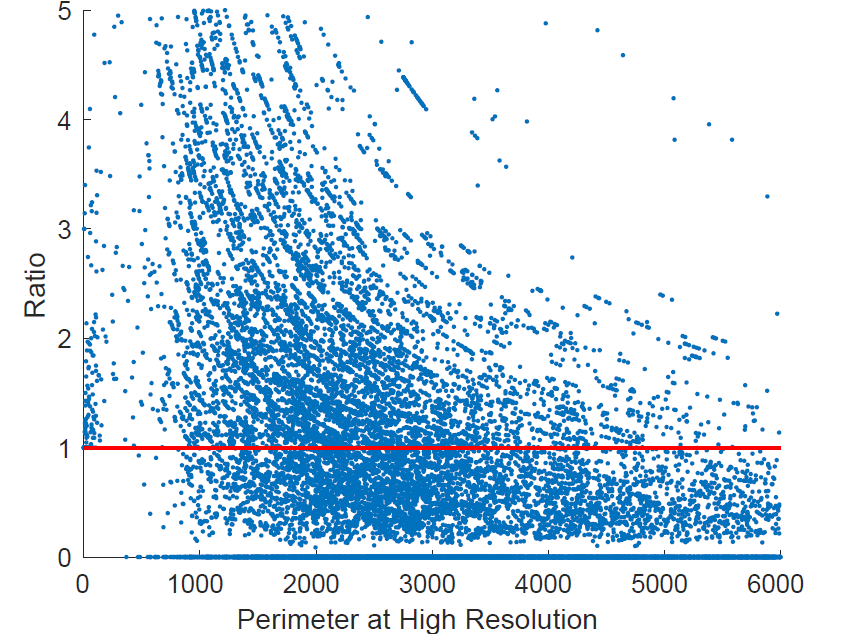} }}%


\caption{Relationship between lake extent size and number of unknown pixels labeled by \orbi (a)}
\label{fig:orbi_scatter}
\end{figure}

Note that even though unknowns in any two extents can be limited to same number of contours but they will have different number of unknowns because of the size of the extents (large extent by definition will have more pixels in its perimeter). We define a metric, $U_{ratio}$ to normalize the number of unknowns with respect to the length of last water contour (perimeter at high resolution) so that the performance at different lake extents can be compared simultaneously. 
\begin{equation}
U_{ratio} = \frac{\#\ unknown\ pixels }{perimeter\ at\ high\ resolution}
\end{equation}
Thus, $U_{ratio}$ can be seen as the approximate measure of number of unknown layers around the true water boundary contour. 
Figure \ref{fig:orbi_scatter} shows the value of $U_{ratio}$ as a function of extent size for 18000 extents derived from 620 lakes for 4 different settings of and $gr$ and $wth$. Note that the probability in Eqn \ref{eq:proof2} is defined with respect to the perimeter at low resolution (C) whereas, in the scatter plots, we have used perimeter at high resolution. This was done to keep the perimeter comparable as the mapping grids are changed. Relationship between probability and perimeter at high resolution can be treat same as with perimeter at low resolution because perimeter at high resolution can be approximated as perimeter at low resolution multiplied by a constant factor. As described in section \ref{sec:orbit_proof}, as the perimeter increases, the probability that unknowns will be localized around the boundary contours increase exponentially which can be empirically observed as exponential decrease in $U_{ratio}$. Since, the probability is inversely proportional to the grid ratio, we observe increase in $U_{ratio}$ (unknows span in more number of layers) when $gr$ is increased from 400 to 1600. On the other hand, as $wth$ is changed from $0.5*gr$ to $0.75*gr$, we observe no change in the relationships which provides empirical evidence that the independence assumption (Assumption A3) stated in \ref{sec:orbit_proof} holds true in real-world situations.

\begin{figure}[!t]
\centering
\subfloat[]{{\includegraphics[width=0.2\textwidth]{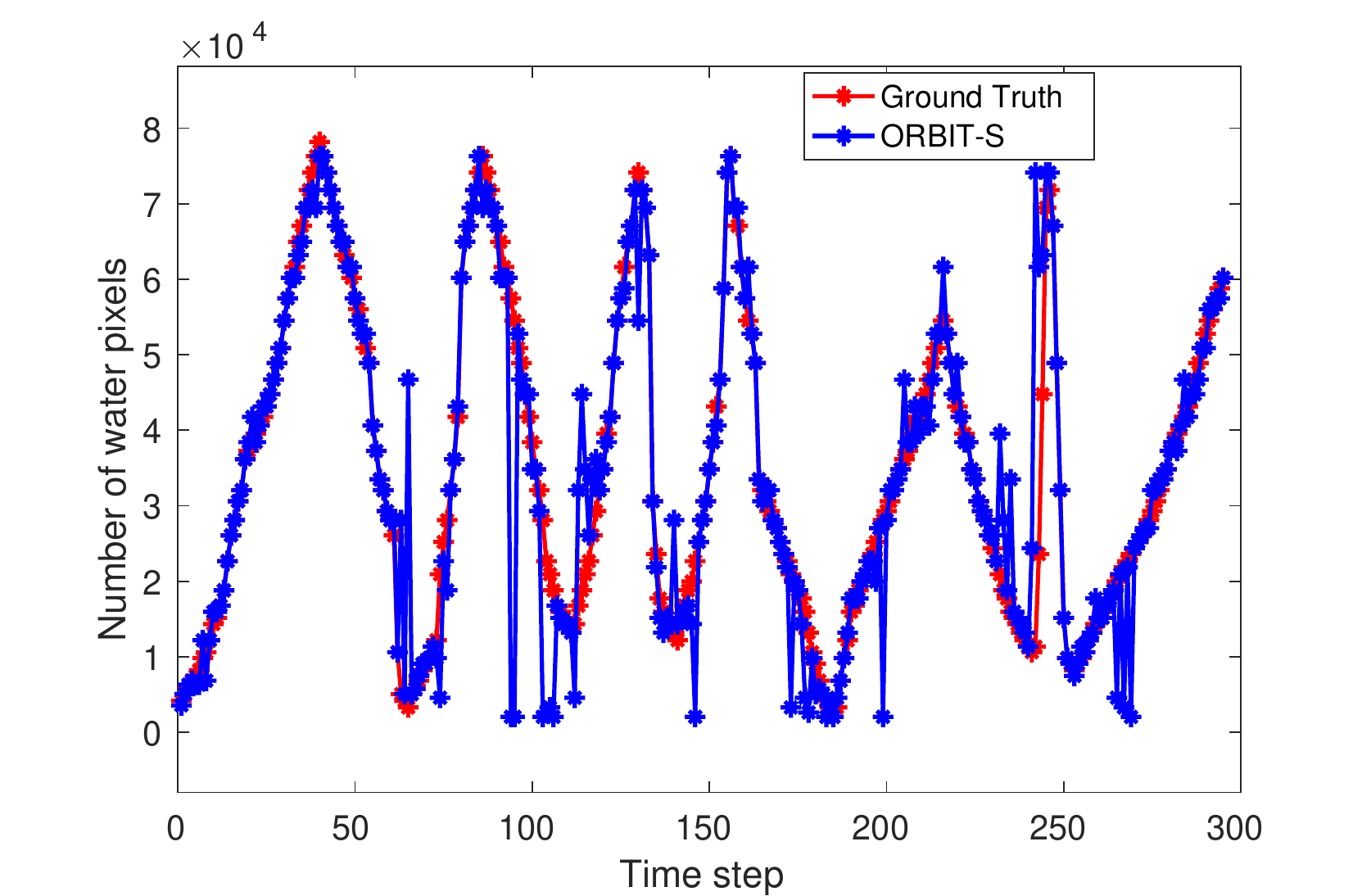}}}%
\subfloat[]{{\includegraphics[width=0.2\textwidth]{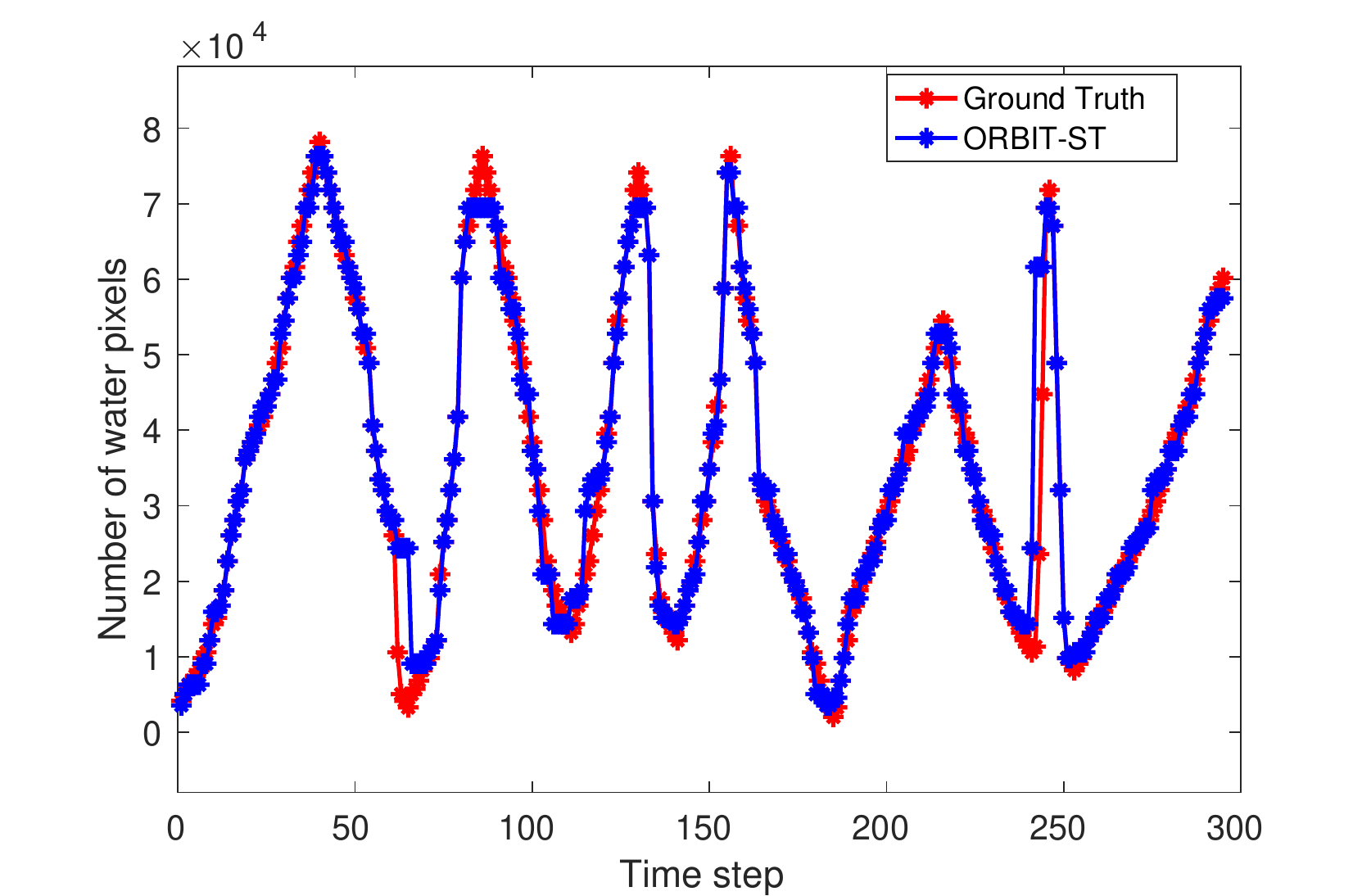} }}%
\caption{Total Area (number of water pixels) time series from both algorithms for $30 \%$ noise case.}
\label{fig:dp_ts}
\end{figure}

\textbf{Impact Of Noise in LSR Labels:}
In the previous experiment we assumed that both LSR labels and high elevation ordering are perfect. The aim of this experiment is to analyze the impact of errors and missing labels in LSR maps on the performance of \orbi\ and \orbst. Here we use elevation structure of Red Fleet Reservoir (available from USGS's 10m Digital Elevation dataset)  and simulate the dynamics (shown in figure \ref{fig:dp_ts}(a) in red color) to create a set of HSR extents. A synthetic square grid of 200m spatial resolution is imposed to simulate the low resolution scale and create corresponding LSR surface extent maps using $wth = 200$. We then added different amounts of spatio-temporally correlated errors and missing labels in LSR maps. Since, the goal of this experiment is to analyze the robustness of \orbi\ and \orbst\ against errors in LSR maps, we use the ground truth HSR ordering available from Digital elevation data. Table \ref{tab:ioe_LSRl} reports the performance of both algorithms for this dataset. First column represents the \% pixels labelled as unknown across all time steps. Second column represents the \% of pixels labelled incorrectly and finally the third column represents total \% of pixels labelled either incorrectly or labelled as unknown. The mean and standard deviation values were generated from 10 random runs of the noise generation process. The values for \orbst\ were generated using $\alpha = 0.8$. As we can see, \orbi\ has labelled very small amount of pixels as unknown and is able to handle large amount of noise in LSR labels even after adding 30 \% noise in LSR labels. \orbst\ further improves the performance by enforcing temporal consistency. Figure \ref{fig:dp_ts} shows the total area time series produced by both algorithms for the 30\% noise case. As we can see, total area time series produced by \orbst\ is significantly smoother. Note that \orbst\ time series is flat around the peak between time steps 50 and 100 which might suggest that the approach is smoothing out the dynamics. However, we don't notice the similar artifact for other peaks which are even narrower which implies that \orbst\ smoothed the peak because those timesteps had a lot of noise (and not because of over-smoothing). To demonstrate the efficacy of the spatio-temporal noise generation process, we applied spatial and temporal majority filters on the noisy LSR maps and report the numbers in table \ref{stab:ioe_LSRl}. Here, we chose the value of $\alpha$ to be 0.8. Next, we provide some discussion on the physical interpretation of $\alpha$ and how to choose it. 

\begin{table}[h]
\centering
\small{
\caption{Impact of Noise in LSR Labels}
\begin{tabular}{|c|c|c|c|}
\hline
  & $\%$ unknown & $\%$ error & $\%$ total \\ \hline
  \multicolumn{4}{c}{\orbi} \\ \hline
5 \% & 0.28 $\pm$ 0.04 & 0.14 $\pm$ 0.14 & 0.42 $\pm$ 0.16 \\ \hline
10 \% & 0.32 $\pm$ 0.02 & 0.33 $\pm$ 0.14 & 0.65 $\pm$ 0.15 \\ \hline
20 \% & 0.66 $\pm$ 0.08 & 1.59 $\pm$ 0.74 & 2.25 $\pm$ 0.75 \\ \hline
30 \% & 0.87 $\pm$ 0.16 & 5.12 $\pm$ 1.12 & 6.00 $\pm$ 1.17 \\ \hline
  \multicolumn{4}{c}{\orbst} \\ \hline
5 \% & 0.19 $\pm$ 0.02 & 0.24 $\pm$ 0.08 & 0.43 $\pm$ 0.07 \\ \hline
10 \% & 0.18 $\pm$ 0.02 & 0.44 $\pm$ 0.15 & 0.61 $\pm$ 0.14 \\ \hline
20 \% & 0.12 $\pm$ 0.03 & 1.21 $\pm$ 0.71 & 1.34 $\pm$ 0.70 \\ \hline
30 \% & 0.11 $\pm$ 0.03 & 3.00 $\pm$ 0.95 & 3.12 $\pm$ 0.95 \\ \hline
\label{tab:ioe_LSRl}
\end{tabular}
}
\end{table}

\begin{table}[h]
\centering
\small{
\caption{Impact of Noise in LSR Labels}
\begin{tabular}{|c|c|c|c|}
\hline
  & $\%$ unknown & $\%$ error & $\%$ total \\ \hline
  \multicolumn{4}{c}{\textbf{Spatial Majority Filtering}} \\ \hline
5 $\%$ & 3.82 $\pm$ 0.76 & 26.22 $\pm$ 0.28 & 26.22 $\pm$ 0.28 \\ \hline
10 $\%$ & 5.14 $\pm$ 0.62 & 29.29 $\pm$ 0.20 & 29.29 $\pm$ 0.20 \\ \hline
20 $\%$ & 8.42 $\pm$ 0.76 & 35.92 $\pm$ 0.39 & 35.92 $\pm$ 0.39 \\ \hline
30 $\%$ & 11.46 $\pm$ 0.62 & 42.38 $\pm$ 0.64 & 42.38 $\pm$ 0.64 \\ \hline
  \multicolumn{4}{c}{\textbf{Temporal Majority Filtering}} \\ \hline
5 $\%$ & 0.74 $\pm$ 0.35 & 4.16 $\pm$ 0.21 & 4.16 $\pm$ 0.21 \\ \hline
10 $\%$ & 1.37 $\pm$ 0.30 & 6.74 $\pm$ 0.30 & 6.74 $\pm$ 0.30 \\ \hline
20 $\%$ & 3.54 $\pm$ 0.41 & 14.10 $\pm$ 0.55 & 14.10 $\pm$ 0.55 \\ \hline
30 $\%$ & 6.53 $\pm$ 0.42 & 23.07 $\pm$ 0.76 & 23.07 $\pm$ 0.76 \\ \hline
\label{stab:ioe_LSRl}
\end{tabular}
}
\end{table}

\textbf{Physical interpretation of $\alpha$:}
\orbt\ optimizes the following objective function to obtain temporally consistent water levels - 

\begin{equation}
\argmin_{\hat{\theta_{1 ... T}}} (Cost_{mismatch}(T) + \alpha Cost_{transition}(T)) 
\label{sup_eq:cost_total}
\end{equation}
and $\alpha \in [0,]$ is the trade-off parameter between two costs. 

\begin{figure}[!t]
\centering
\subfloat[]{{\includegraphics[width=0.2\textwidth]{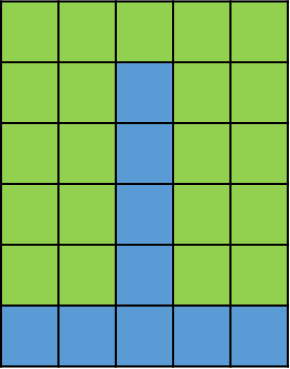}}}%
\hfill
\subfloat[]{{\includegraphics[width=0.2\textwidth]{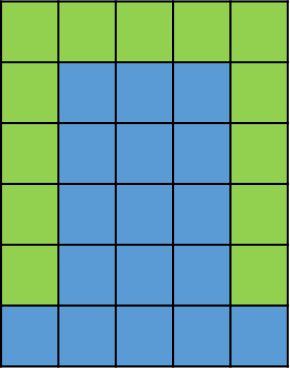} }}%
\qquad
\caption{Relationship between $\alpha$ and persistence of area dynamics. (a) $Cost_{transition} = 8$, $Cost_{mismatch} = 4$ (b) $Cost_{transition} = 8$, $Cost_{mismatch} = 12$}
\label{fig:alpha2}
\end{figure}

As $\alpha$ is increased, the temporal smoothness in water level variations will increase to reduce $Cost_{transition}$. However, over-smoothing of water level variations would not only remove spurious changes but also some real dynamics. In particular, if the cost of removing the dynamics ($Cost_{mismatch}$) is less than the $\alpha*Cost_{transition}$, then the dynamics will be removed. For example,  consider the two different dynamics shown in figure \ref{fig:alpha2}. Dynamics in figure \ref{fig:alpha2} (a) represents the most abrupt dynamics that can be present in our data. This dynamics will get removed for $\alpha >0.5$. For figure \ref{fig:alpha2} (b), $\alpha>1.5$ would remove the dynamics. 

\begin{figure}[!t]
\centering
\subfloat[]{{\includegraphics[width=0.24\textwidth]{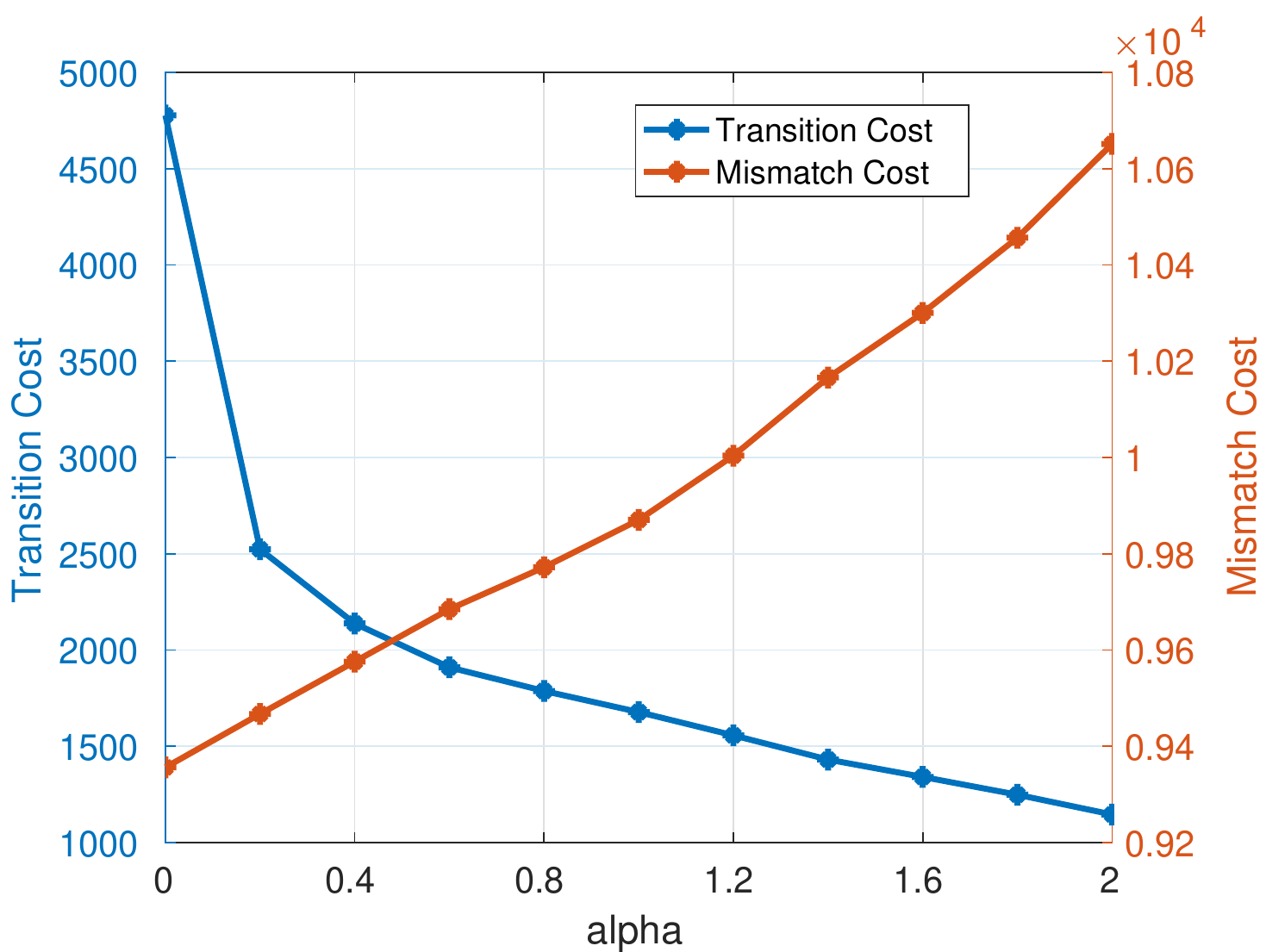}}}%
\subfloat[]{{\includegraphics[width=0.24\textwidth]{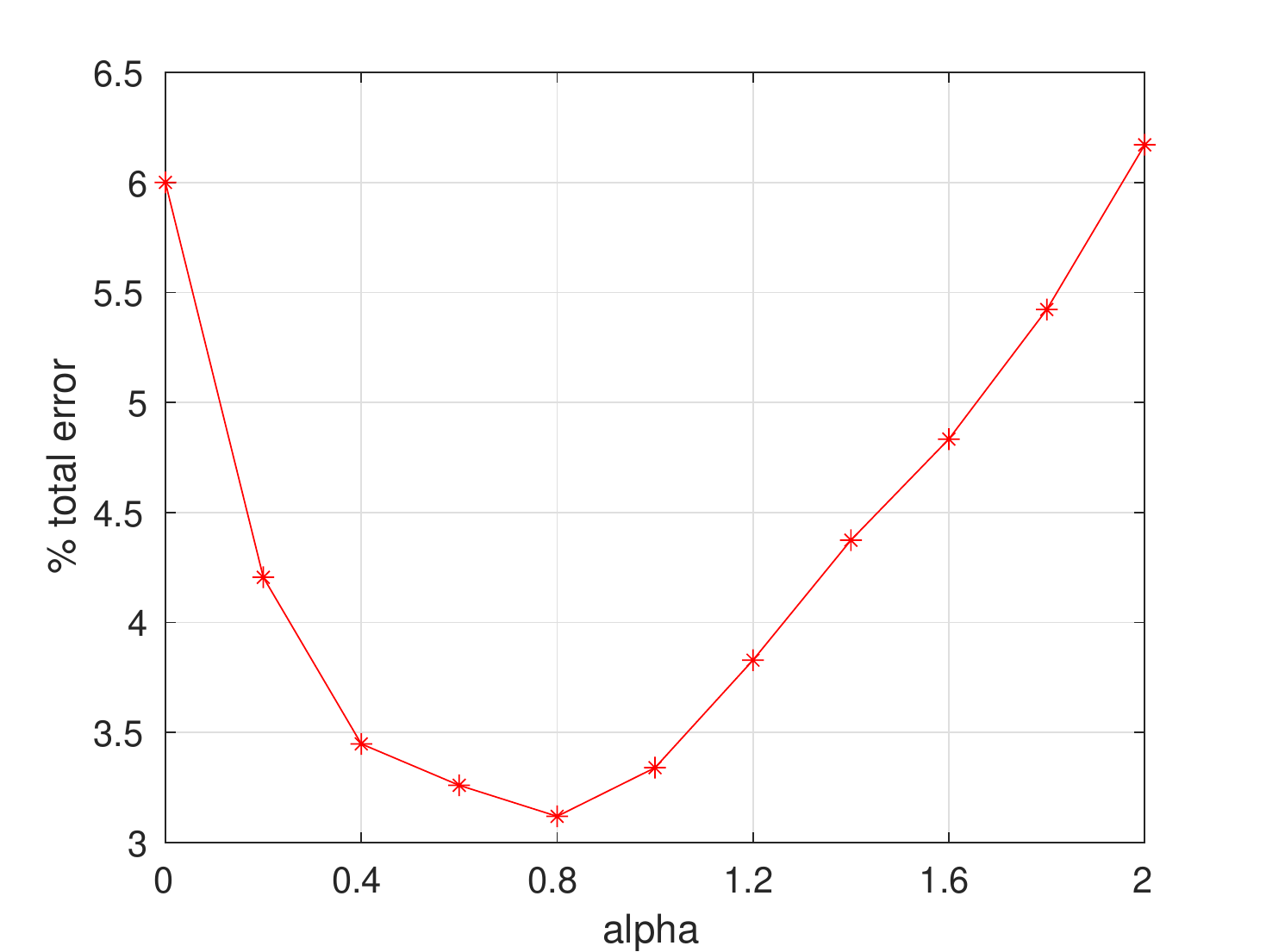} }}%
\qquad
\caption{Impact of $\alpha$ on performance for the 30 \% noise case. (a) Variation of $Cost_{mismatch}$ and $Cost_{transition}$ as a function of $\alpha$ (b) \% total error as a function of $\alpha$}
\label{fig:alpha}
\end{figure}

The relationship between $Cost_{mismatch}$ and $Cost_{transition}$ can be used to guide the search for the right value of $\alpha$. $Cost_{mismatch}$ and $Cost_{transition}$ are inversely related i.e. as $\alpha$ is increased, $Cost_{mismatch}$ decreases and $Cost_{transition}$ increases. However, both these costs change at different rates. Specifically, if there are spurious changes in total area values, removing them would lead to large drop in $Cost_{transition}$ without increasing $Cost_{mismatch}$ significantly as illustrated in figure \ref{fig:alpha}. On the other hand if the change in area values is not spurious then both cost will change slowly. Hence, $Cost_{transition}$ and $Cost_{mismatch}$ curves can be used to decide the value of $\alpha$. Figure \ref{fig:alpha}(a) shows these curves for 30 \% noise case. As we can see, $Cost_{transition}$ is decreasing rapidly till $\alpha = 0.6$ and then it decreases linearly. Hence, $\alpha$ should be chosen to be close to 0.6. Figure \ref{fig:alpha}(b) shows the \% total error as the function of $\alpha$. As we can see, the performance at $\alpha=0.6$ is very close to the best performance (achieved at $\alpha = 0.8$).

\subsection{Real Data Experiments}\hfill

\begin{figure}[!t]
\centering
\subfloat[]{\frame{\includegraphics[width=0.2\textwidth]{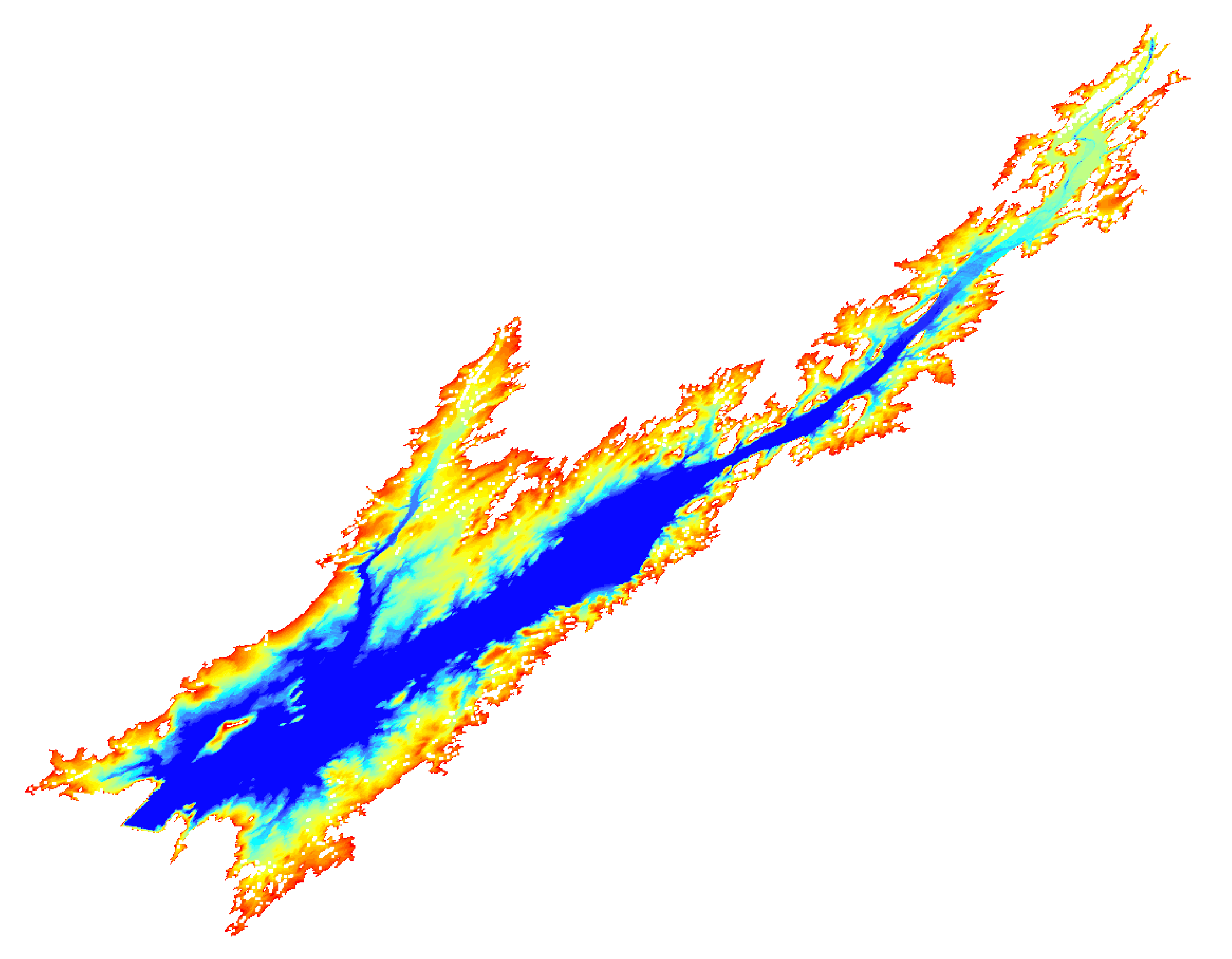}}}
\hfill
\subfloat[]{\frame{\includegraphics[width=0.2\textwidth]{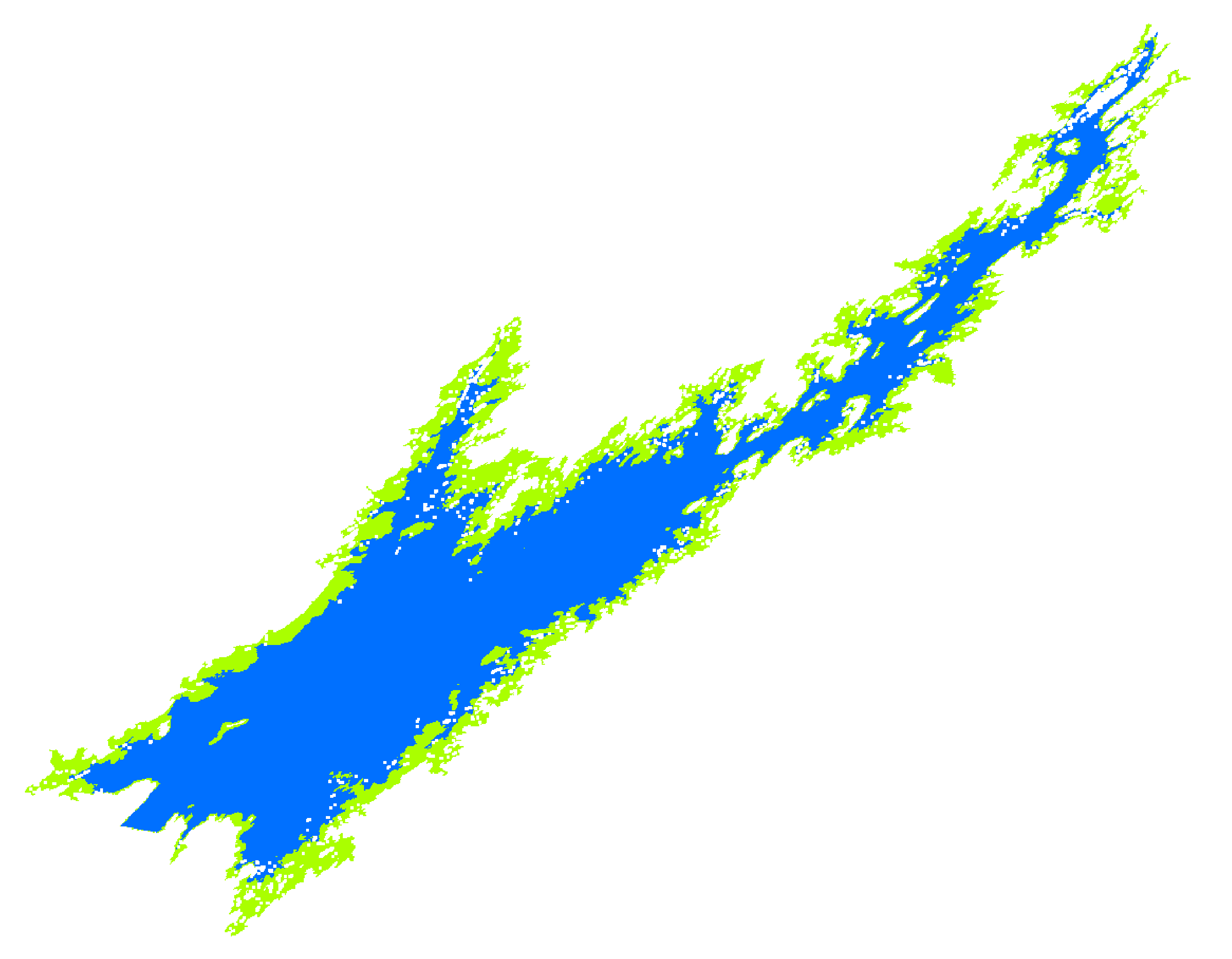} }}%

\subfloat[]{\frame{\includegraphics[width=0.15\textwidth]{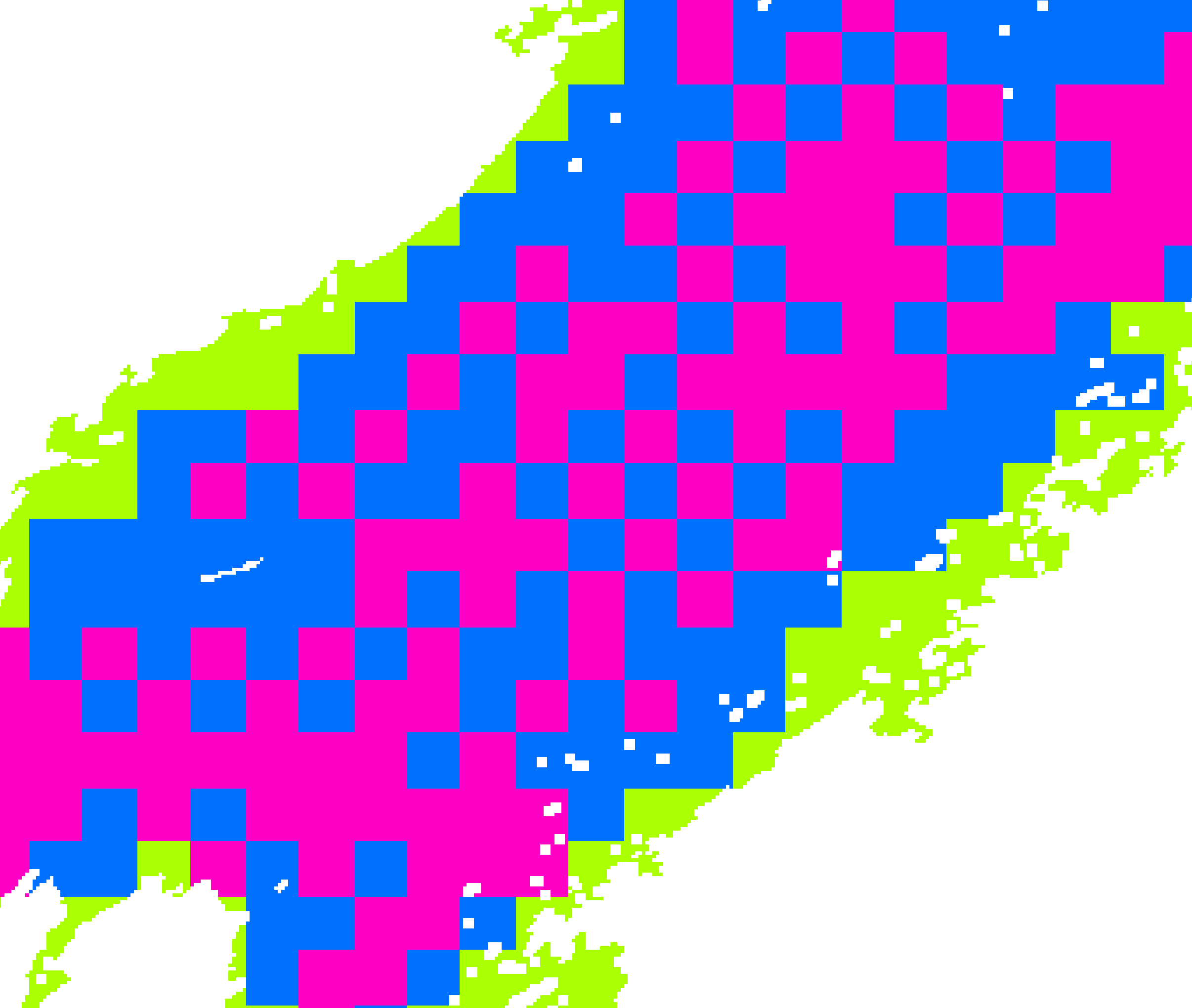} }}%
\hfill
\subfloat[]{\frame{\includegraphics[width=0.15\textwidth]{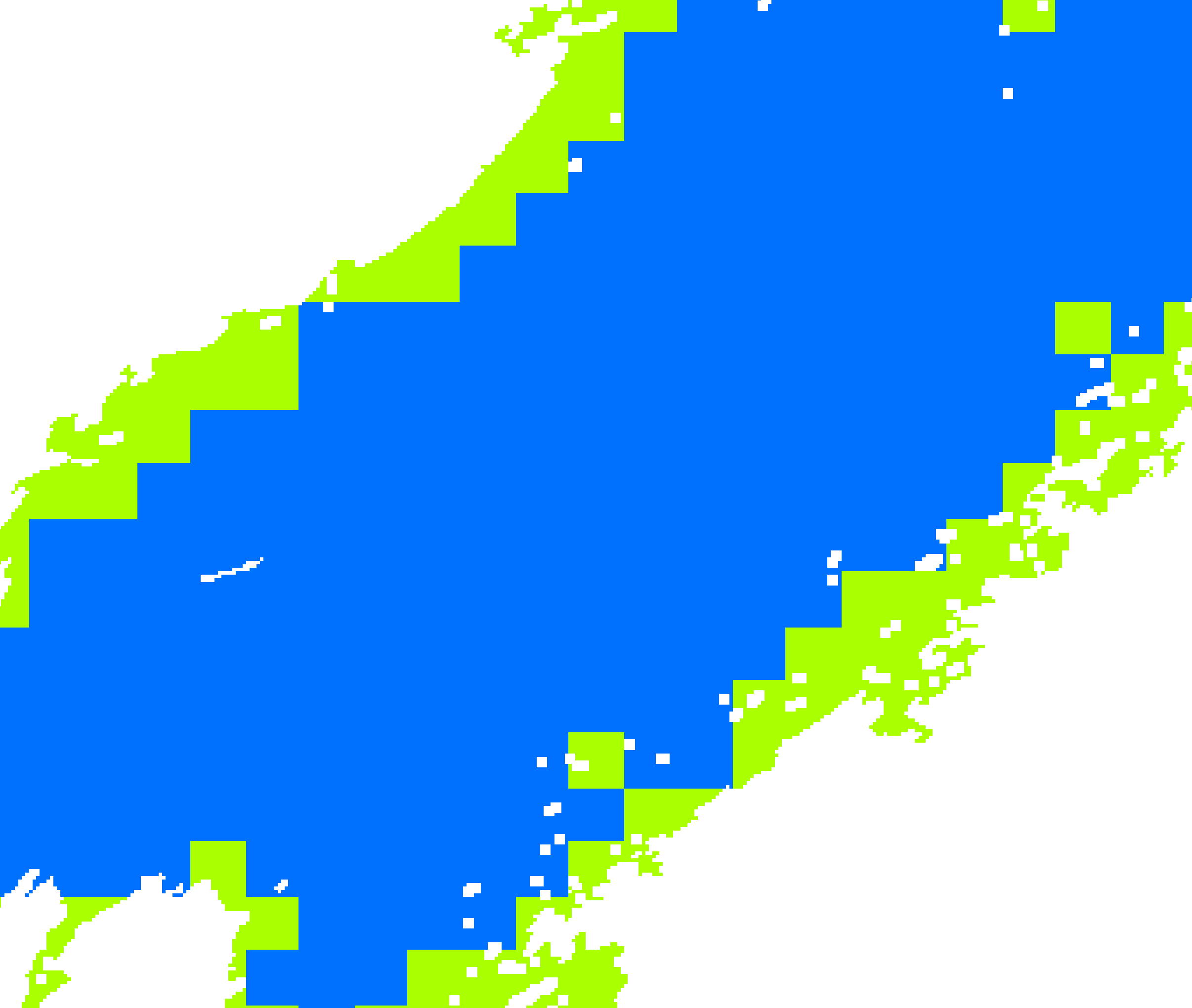} }}%
\hfill
\subfloat[]{\frame{\includegraphics[width=0.15\textwidth]{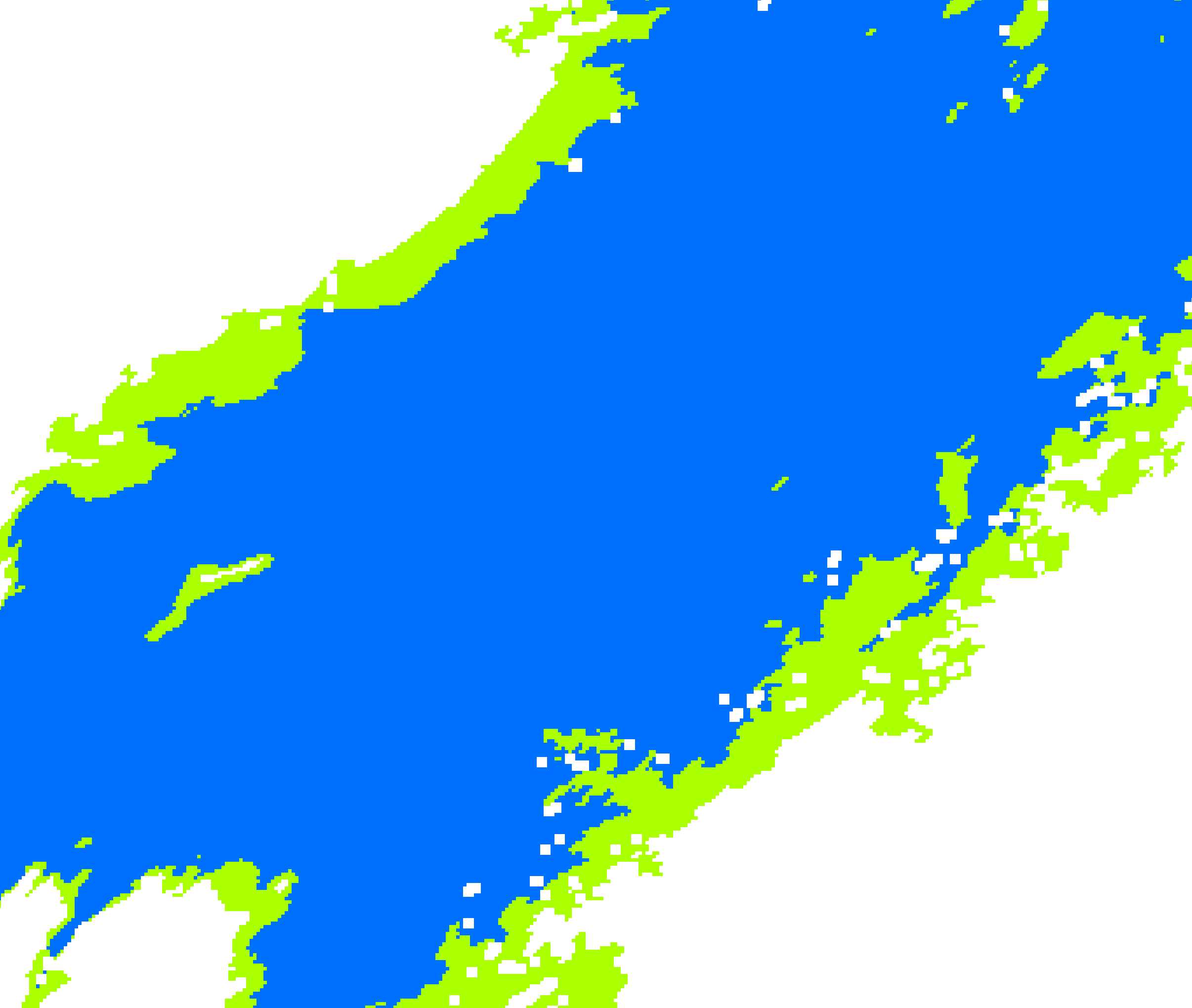} }}%

\subfloat[]{\frame{\includegraphics[width=0.2\textwidth]{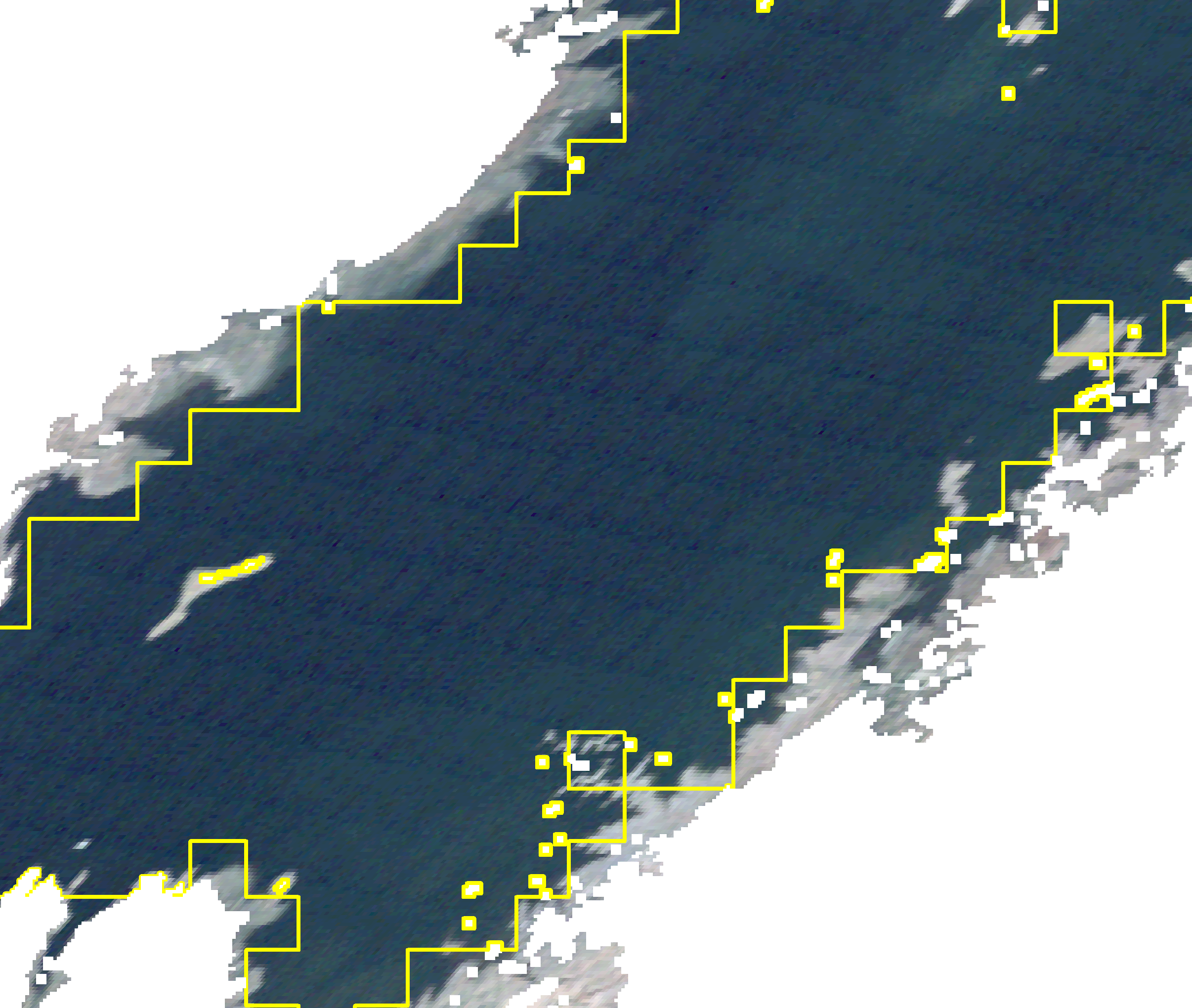}}}
\hfill
\subfloat[]{\frame{\includegraphics[width=0.2\textwidth]{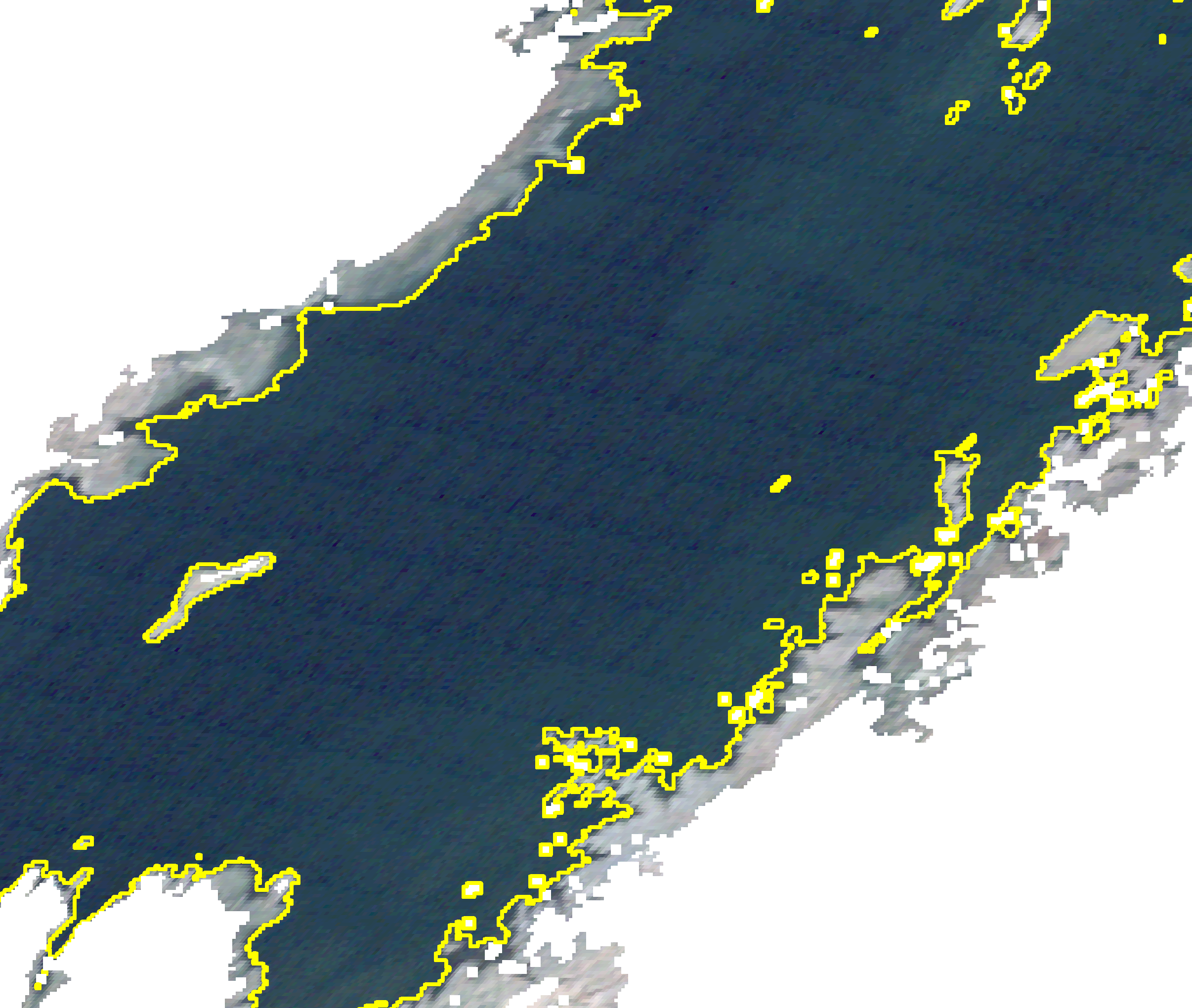} }}%

\caption{Performance of ORBIT framework for Lake Oviachic, Mexico on Oct 16, 2006. (a) Relative Elevation Ordering (b) Ground Truth (c) Input LSR extent (d)Corrected LSR extent (e) Estimated HSR extent (f) Corrected LSR extent boundary (g) Estimated HSR extent boundary}
\label{fig:real-data-exp}
\end{figure}

\textbf{Impact of Noise in LSR and HSR labels:} In this experiment, we will assume that we don't have access to neither perfect elevation ordering nor perfect LSR labels. Here, we used noisy multi-temporal classification maps at LANDSAT scale (30m) from JRC product to learn relative elevation ordering at HSR. We obtained noisy daily multi-temporal classification maps using methodology described in \cite{khandelwal2017approach} on MODIS daily multispectral data at 500m (MOD09GA product \cite{lpdaac}). We then applied \orbst\ to obtain HSR extent maps at daily scale. Figure \ref{fig:real-data-exp} demonstrate the effectiveness of \orbi\ paradigm for Lake Oviachic in Mexico on Oct 16, 2006. Figure \ref{fig:real-data-exp} (a) shows the relative ordering learned from JRC data. In order to easily visualize spatial patterns, we show only the bottom part of the lake in Figures \ref{fig:real-data-exp} (c)-(g). Figure \ref{fig:real-data-exp} (c) shows the noisy classification map at LSR. As we can see there are lot of missing pixels as they were filtered out due to poor data quality. Figure \ref{fig:real-data-exp} (d) shows the corrected classification map at LSR where labels of missing pixels were successfully recover. Also note the blocky nature of classification maps due to low spatial resolution of MODIS data. Figure \ref{fig:real-data-exp} (e) show the estimated HSR classification map. The estimated HSR map much better spatial detail. In Figure \ref{fig:real-data-exp} (f) and (g) we show the estimated LSR and HSR extents overlaid on LANDSAT true color composite image on Oct 16, 2006. As we can see, estimated HSR extent matches very well water boundary as seen from the LANDSAT data.

\section{Conclusions and Future Work}
\label{sec:future}
In this paper, we proposed a new framework, ORBIT that uses relative elevation ordering among instances to effectively transfer information between complementary spatio-temporal datasets. We provided insights into different aspects of the framework using both synthetic and real-world dataset for the of global surface water monitoring.

For future work, we aim to advance the algorithms from several aspects. In this paper, we assumed a single cut-off threshold ($wth$) for all pixels and timesteps. We aim to develop iterative methods to estimate $wth$ separately for each pixel using its multi-temporal class label information. 
Currently, the step for obtaining ordering at HSR does not incorporate any information from LSR multi-temporal maps. We aim to extend \orbi\ to make use of information from LSR extent maps to improve the quality of elevation ordering. The key idea is that if the errors in two datasets are not correlated to each other then bringing information from LSR extent maps can improve ordering quality.
Using high quality elevation structure information for a wide variety of water bodies, we showed that the performance of the framework depends on the extent perimeter. In future, we aim to make the framework more effective for smaller lakes by making use of noisy fractional labels rather than binary labels. \orbc\ approach can be extended to detect changes in the elevation structure by using moving window based strategy, where instead of using the whole duration to learn elevation ordering, it is learned for a given window and then utilized to find the locations that consistently disagree with the current ordering. Finally, even though both existing DEM datasets and elevation ordering learned from data can have errors, they can be merged together to get a better overall ordering. 

\bibliographystyle{IEEEtran}
\bibliography{bare_conf.bib}
\end{document}